\documentclass[runningheads]{llncs}

\usepackage{eccv}

\usepackage{multirow}
\usepackage{makecell}
\usepackage{nicefrac}

\usepackage{eccvabbrv}

\usepackage{graphicx}
\usepackage{booktabs}
\usepackage{caption}

\DeclareMathOperator*{\argmin}{arg\,min}

\usepackage{subcaption}
\usepackage{wrapfig}
\usepackage{multirow}
\newcommand{\mathbold}[1]{{\boldsymbol{{#1}}}}

\newcommand{\nestedmathbold}[1]{{\mathbold{#1}}}

\newcommand{\mbc}{\nestedmathbold{c}}

\newcommand{\mbm}{\nestedmathbold{m}}

\newcommand{\mbr}{\nestedmathbold{r}}
\newcommand{\mbs}{\nestedmathbold{s}}

\newcommand{\mbu}{\nestedmathbold{u}}

\newcommand{\mbw}{\nestedmathbold{w}}
\newcommand{\mbx}{\nestedmathbold{x}}

\newcommand{\mbz}{\nestedmathbold{z}}

\newcommand{\mbepsilon}{\nestedmathbold{\epsilon}}

\newcommand{\mbmu}{\nestedmathbold{\mu}}

\newcommand{\mbSigma}{\nestedmathbold{\Sigma}}

\newcommand{\cJ}{\mathcal{J}}

\usepackage{algorithmic}
\usepackage[ruled]{algorithm2e}

\usepackage{indentfirst}
\usepackage[accsupp]{axessibility}  

\usepackage[colorlinks]{hyperref}

\usepackage{orcidlink}
\definecolor{MyCyan}{RGB}{0,163,218}
\definecolor{MyDarkBlue}{RGB}{0,103,165}
\definecolor{MyDarkGreen}{RGB}{56,116,51}
\definecolor{MyMagenta}{RGB}{200,18,126}

\makeatletter
  \newcommand\figcaption{\def\@captype{figure}\caption}
  \newcommand\tabcaption{\def\@captype{table}\caption}
\makeatother

\begin{document}

\title{Optimizing Diffusion Models for Joint Trajectory Prediction and Controllable Generation} 

\titlerunning{OptTrajDiff}

\author{Yixiao Wang\inst{1} \and
Chen Tang\inst{1,2} \and
Lingfeng Sun\inst{1} \and
Simone Rossi\inst{3} \and
Yichen Xie\inst{1}\and
Chensheng Peng\inst{1}\and
Thomas Hannagan\inst{3}\and
Stefano Sabatini\inst{3}\and
Nicola Poerio\inst{4}\and
Masayoshi Tomizuka\inst{1}\and
Wei Zhan\inst{1}}

\authorrunning{Y. Wang et al.}

\institute{University of California, Berkeley, USA \and The University of Texas at Austin, USA \and Stellantis, France \and Stellantis, Italy \\
\email{\{yixiao\_wang, wzhan\}@berkeley.edu}
}

\maketitle

\begin{abstract}
 Diffusion models are promising for joint trajectory prediction and controllable generation in autonomous driving, but they face challenges of inefficient inference steps and high computational demands. To tackle these challenges, we introduce Optimal Gaussian Diffusion (OGD) and Estimated Clean Manifold (ECM) Guidance. OGD optimizes the prior distribution for a small diffusion time $T$ and starts the reverse diffusion process from it. ECM directly injects guidance gradients to the estimated clean manifold, eliminating extensive gradient backpropagation throughout the network. Our methodology streamlines the generative process, enabling practical applications with reduced computational overhead. Experimental validation on the large-scale Argoverse 2 dataset demonstrates our approach's superior performance, offering a viable solution for computationally efficient, high-quality joint trajectory prediction and controllable generation for autonomous driving. Our project webpage is at \url{https://yixiaowang7.github.io/OptTrajDiff_Page/}
\end{abstract}


\section{Introduction}
\label{sec:intro}
The diffusion model is a class of generative models capable of representing high-dimensional data distributions. In particular, it has demonstrated strong performance in trajectory prediction and generation for autonomous driving~\cite{gu2022stochastic,zhong2023guided, rempe2023trace, jiang2023motiondiffuser, guo2023scenedm}. In contrast to traditional trajectory prediction \cite{zhou2023query, lan2023sept} and generative models\cite{suo2021trafficsim,xu2023bits}, the unique advantage of diffusion models lies in their ability to deform the generated trajectory distribution to comply with additional requirements at inference stage via gradient-based guided sampling. Notably, it is achieved without extra model training costs. This ability to perform controllable trajectory generation enables various useful applications, such as enforcing additional realism constraints on predicted trajectories, generating directed and user-specified simulation scenarios. 

However, computational efficiency is a crucial bottleneck hindering the practical application of diffusion models in autonomous driving. Real-time inference is essential for trajectory prediction, as it provides timely forecasts of surrounding agents' behavior, enabling safe and efficient planning in dynamic traffic scenarios. The high demand for inference speed, coupled with limited onboard computational resources, makes it infeasible to deploy diffusion models for onboard trajectory prediction. While simulations do not occur onboard, lightweight models are still preferred to streamline the closed-loop training and evaluation pipelines. The heavy computational cost is mainly attributed to the following two aspects:

\textbf{Computationally Expensive Reverse Diffusion.} At the inference stage, the diffusion model samples from standard Gaussian distribution and gradually denoises the sample through dynamics described by a Stochastic Differential Equation (SDE) \cite{song2020score}, aiming to eventually obtain a sample as if drawn from a target data distribution. The target data distribution can be significantly different from a standard Gaussian distribution, necessitating a large number of denoising steps to yield good performance. Prior works have attempted to reduce the reverse diffusion steps through adaptive noise schedule~\cite{san2021noise, kingma2021variational}, fast samplers~\cite{song2020denoising, lu2022dpm,watson2022learning,zhang2022fast}, or distillation~\cite{salimans2022progressive,song2023consistency}. However, the fixed standard Gaussian prior poses a challenge in accelerating the reverse diffusion process without violating the SDE, which can inevitably compromise the generation quality.

\textbf{Computationally Expensive Guided Sampling.} Controllable generation is typically achieved by guiding the denoising process with the gradient of a guidance cost function. The guidance cost function encodes the desired characteristics of the generated data, which is typically defined on the clean data manifold in trajectory prediction and controllable generation problems. However, guided sampling intends to inject the gradient of the guidance cost function into the series of noisy data at intermediate diffusion steps. It requires a forward pass to estimate the clean data first and then back-propagating throughout the entire network to estimate the gradient with respect to the intermediate noisy data~\cite{rempe2023trace,jiang2023motiondiffuser,lin2024joint}, which is very computationally intensive.

Targeted at these challenges, we take a step further to improve the computational efficiency of diffusion models while maintaining their performance for joint trajectory prediction and controllable generation tasks. Specifically, we propose two novel solutions for efficient reverse diffusion and guided sampling respectively. First, we present \textit{Optimal Gaussian Diffusion} (OGD) to accelerate the reverse diffusion process. At the inference stage, instead of a standard Gaussian distribution far away from the desired data distribution, OGD starts from an optimal Gaussian distribution, which minimizes the distance to intermediate data distribution at a specific noise level, cutting down the diffusion steps before that. We show that we can analytically estimate such an optimal Gaussian distribution and also, an optimal perturbation kernel distribution, \emph{at any noise level} from the statistics of the data distribution. It allows flexible tuning of the diffusion steps at the inference stage, without the need to train any additional models~\cite{zheng2022truncated}. We further derive a practical implementation of OGD for joint trajectory prediction and generation, where the optimal Gaussian prior is computed using the mean and variance of marginal trajectory distributions estimated with a pre-trained marginal trajectory prediction~\cite{nayakanti2023wayformer,zhou2023query, lan2023sept} model.

Second, we propose \textit{Estimated Clean Manifold Guidance} (ECM) to accelerate guided sampling for controllable generation. To save the computational cost due to estimating the guided gradient on the noisy data, we aim to directly inject the gradient into the clean data manifold without lengthy backpropagation. ECM is motivated by the insight that guided sampling can be regarded as a multi-objective optimization problem on the clean data manifold: The first objective is to maximize the likelihood of the samples on the estimated real data distribution; the second objective is to achieve low guidance cost. ECM hierarchically solves this multi-objective problem without backpropagation throughout the entire diffusion model. We show that it leads to faster inference time and much better performance than existing approaches.  Also, to tackle the challenges imposed by the multi-modal nature of vehicle interactions, we propose to warm-start the multi-objective optimization problem with reference joint trajectory points estimated using a marginal trajectory predictor. We refer to the complete algorithm as \emph{Estimated Clean Manifold with Reference Joint Trajectory} (ECMR). 

To evaluate the proposed OGD and ECM methods on real-world tasks, we implement the OGD model leveraging a pre-trained marginal prediction model, QCNet~\cite{zhou2023query}, and conduct extensive experiments on the Argoverse 2 dataset. We show that OGD can achieve better joint trajectory prediction performance than vanilla diffusion with significantly fewer diffusion steps\textemdash it only takes \emph{$\nicefrac{1}{12}$} of the diffusion steps used by the vanilla diffusion model. OGD also achieves outstanding prediction accuracy compared to non-diffusion joint prediction models. In addition, ECMR, coupled with OGD, can generate samples with significantly lower guidance costs with the same level of realism scores compared to conducting controllable generation on the vanilla diffusion model using existing guided sampling approaches used in autonomous driving \cite{rempe2023trace, jiang2023motiondiffuser}, but using around \emph{$\nicefrac{1}{5}$} of their inference step. 

\begin{figure}[t]
    \centering
    \begin{subfigure}[b]{0.5\textwidth}
            \includegraphics[width=\linewidth]{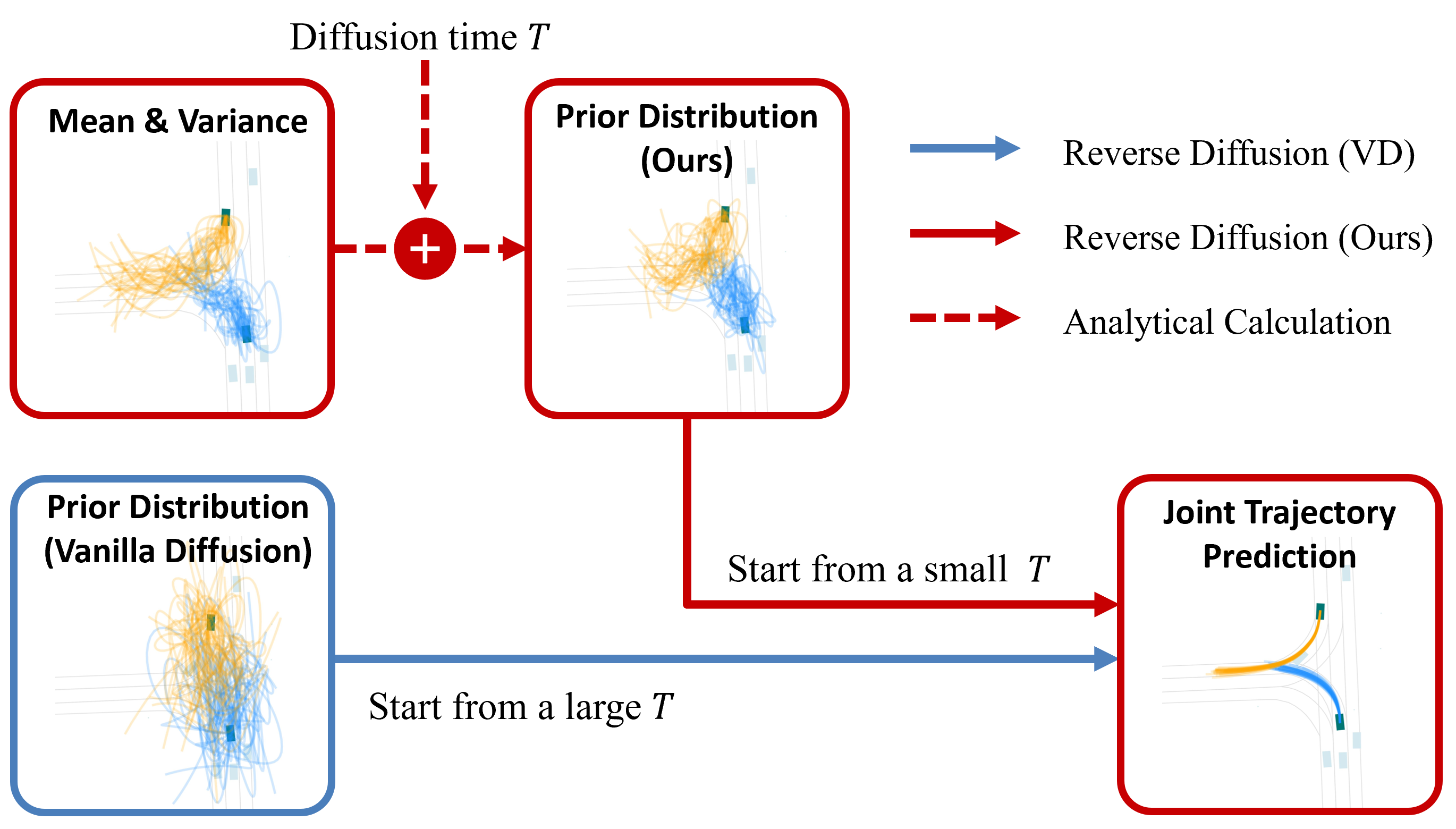}
            \caption{Optimal Gaussian Diffusion (OGD)}
            \label{fig:ogd_vd}
    \end{subfigure}%
    \begin{subfigure}[b]{0.5\textwidth}
            \includegraphics[width=\linewidth]{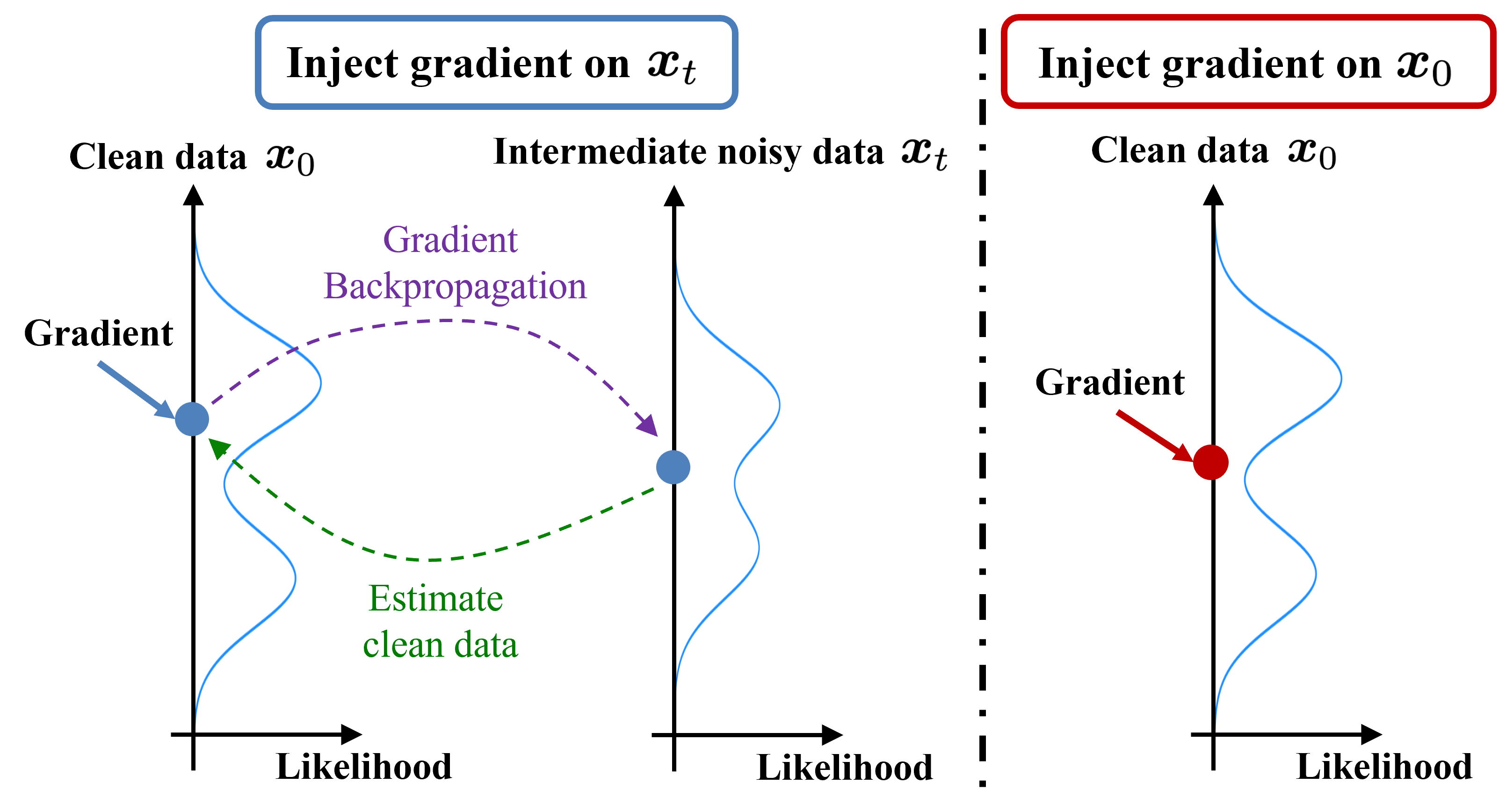}
            \caption{Estimated Clean Manifold Guidance (ECM)}
            \label{fig:ecm}
    \end{subfigure}%
    
    \caption{Overview of Optimal Gaussian Diffusion (OGD) and Estimated Clean Manifold Guidance (ECM). {\bf (a)} OGD uses the mean and variance of the data distribution to calculate the optimal prior distribution at a small $T$. It can largely reduce the diffusion time compared with vanilla diffusion. {\bf (b)} ECM directly injects the gradient of guidance into the clean data manifold to mitigate computational complexity.}\label{fig:controll goal point}
\end{figure}

\section{Related Work}
\label{sec:related work}
{\bf Diffusion models.} Diffusion models have demonstrated their ability to produce high-quality, diverse samples in a variety of applications, such as image, video, and 3D generation \cite{rombach2022high, ho2022video, luo2021diffusion}. Recently, diffusion models have been applied to trajectory prediction in autonomous driving. It shows great performance on representing future trajectory distribution \cite{gu2022stochastic, jiang2023motiondiffuser}. However, diffusion models need to run lengthy reverse diffusion processes to generate high-quality samples \cite{gu2022stochastic, franzese2023much}. This makes it hard to apply diffusion models to trajectory prediction in autonomous driving since it requires in-time prediction for the downstream planning module to promptly respond to dynamically changing traffic scenarios. Previous works~\cite{san2021noise, kingma2021variational, song2020denoising, lu2022dpm,watson2022learning,zhang2022fast, salimans2022progressive,song2023consistency} mostly focused on mitigating this issue by investigating how to solve reverse SDE in a faster manner. In addition, similar to ours, some works sought alternative initialization of the reverse diffusion process to achieve faster inference. For example, \cite{chung2022come} initialized the reverse diffusion process with samples generated by another generative backbone network. However, the backbone network is not deliberately trained for the reverse diffusion process. \cite{zheng2022truncated} proposed to learn the initial diffusion with a generative adversarial network (GAN)~\cite{goodfellow2014generative}. However, it requires training a large additional model with a complex training procedure. Also, it requires specifying the reverse diffusion time as a hyperparameter prior to GAN training, which is hard to tune. 

{\bf  Guided sampling.} 
Diffusion models have been successfully used to tackle controllable tasks through guided sampling, such as image inpainting \cite{song2022pseudoinverse} and motion planning \cite{janner2022planning}. A notable feature of diffusion models with guided sampling is their ability to condition the generation process on the user's preference, which was not available during the training phase. In the driving domain, recent works used guided sampling to generate controllable traffic~\cite{zhong2023guided, jiang2023motiondiffuser}, and controllable pedestrian animation~\cite{rempe2023trace}. Our work belongs to the prior category, where user-specified guidance cost is used to guide generation in the trajectory space. In this case, the guidance cost function encodes certain desired properties of the generated trajectories. Thus, it is normally defined on the realistic trajectory samples, which are on the so-called clean manifold, instead of the noisy manifold containing the intermediate noisy data. Some works attempted to learn the guidance cost function on the noisy manifold \cite{janner2022planning}. Otherwise, it will lead to numerical instability when evaluating guidance cost at intermediate noisy data \cite{zhong2023guided, rempe2023trace}. To avoid the additional computational costs introduced by a learned guidance cost function, \cite{jiang2023motiondiffuser, rempe2023trace} proposed to project the intermediate noisy data into the clean manifold through the diffusion model, and evaluate the guidance cost on the projected point. This approach requires back-propagating throughout the entire diffusion model, which is also computationally intensive. 

{\bf Trajectory prediction.}
In autonomous driving \cite{peng2024q, peng2023delflow}, it is vital to precisely forecast how other participants in traffic will move in the future so that ego vehicle can plan a safe and efficient trajectory to execute in the future. Marginal trajectory prediction is used to predict the trajectory distribution for single vehicle and recent works involve kinematic constraints of the vehicles, restrictions of complex topology of roads, and interaction from the surrounding vehicles \cite{nayakanti2023wayformer,zhou2023query, lan2023sept}. Recently, joint trajectory prediction has attracted the attention of researchers. It consists in predicting the joint future trajectories for all agents so that these trajectories are consistent with one another \cite{ngiam2021scene, xu2022pretram, sun2022pseudolabel, yin2021diverseinteraction}, an aspect which marginal trajectory prediction does not consider. Scene-Transformer \cite{ngiam2021scene} uses a fixed set of learnable scene embeddings to generate corresponding joint trajectories for all the vehicles in the given scene. Models like M2I \cite{sun2022m2i}, and FJMP \cite{rowe2023fjmp} adopt a conditional approach, predicting the motions of other agents based on the movements of controlled agents. Diffusion model \cite{jiang2023motiondiffuser} has also been used to predict the joint trajectory. However, joint trajectory prediction is still a challenging problem since the complexity increases exponentially with the number of vehicles in the scene. The efficiency problem becomes more serious when using diffusion model to predict the joint trajectory distribution \cite{jiang2023motiondiffuser}, and largely limits the application of diffusion models in autonomous driving. 

\section{Preliminaries}
\subsection{Diffusion Models}
\label{sec: pre diffusion}
The diffusion process continuously perturbs the unknown data distribution $p_{data}$ with a perturbation kernel and generates intermediate data with a given diffusion time $T$. Denote the distribution of the time-dependent intermediate nosiy data $\mbx_t$ as $p_t(\mbx_t)$, $t \in [0,T]$, where $p_0(\mbx_0)=p_{data}$ is the clean data distribution and $\mbx_0$ is the clean data. The series of intermediate data $\mbx_t$ are generated through the Stochastic Differential Equation (SDE) \cite{song2020score}:
\begin{equation}
    \text{d}\mbx_t=f(\mbx_t,t)\text{d} t+g(t) \text{d} \mbw , \mbx_0 \sim p_{data} = p_0(\mbx_0),
\end{equation}
where $f(\cdot, t)$ is the drift coefficient, $g(t)$ is the diffusion coefficient, and $\mbw$ is the Wiener process. We can recover $p_{data}$ from reverse-time SDE
\begin{equation}
\label{eq: reverse sde}
    \text{d}\mbx_t=[f(\mbx_t,t)-g(t)^2 \nabla_{\mbx_t} \log p_t(\mbx_t) ]\text
{d}t+g(t)\text{d}\bar{\mbw}, \mbx_T \sim p_T(\mbx_T), 
\end{equation}
where $\bar{\mbw}$ is another Wiener process, $\text{d} t$ is negative timestep. 

In order to solve \cref{eq: reverse sde}, we first need to approximate $p_T(\mbx_T)$. In previous works, $p_T(\mbx_T)$ is typically approximated by some prior distributions $p_{prior}$ which contain no information of $p_{data}$. In this paper, we adopt the setting of Variance Preserving (VP) SDE~\cite{ho2020denoising, song2020score}. In VP-SDE, $p_T(\mbx_T) \approx p_{prior}$ when $T \rightarrow \infty$. The perturbation kernels are in the form of $p_t(\mbx_t|\mbx_0) = \mathcal{N}(\mbx_t;\sqrt{\Bar{\alpha}_t} \mbx_0, (1-\Bar{\alpha}_t)\mbSigma_p)$ where scalar $\Bar{\alpha}_t$ is diffusion schedule parameter, $|\mbSigma_p|=1$. A common choice for $\mbSigma_p$ is the identity matrix $\mathbf{I}$ \cite{ho2020denoising, song2020score}. 
Second, we need to approximate $\nabla_{\mbx_t} \log p_t(\mbx_t)$ for all $t\in(0, T]$. Some works solve score-matching problem \cite{hyvarinen2005estimation, song2019generative} and learn a score function $\mbs_\theta(\mbs_t,t)$ to approximate $\nabla_{\mbx_t} \log p_t(\mbx_t)$. In this paper, we follow the practice in DDPM~\cite{ho2020denoising} 
to learn the noise $\mbepsilon$ using a network $\mbepsilon_\theta(\mbx_t,t)$:
\begin{equation}
    \argmin_\theta \mathbb{E}_{\mbx_0 \sim p_{data}, \mbepsilon \sim \mathcal{N}(\mathbf{0},\mbSigma_p)}||\mbepsilon-\mbepsilon_\theta(\mbx_t,t)||_2^2
\end{equation}
where $\mbx_t = \sqrt{\Bar{\alpha}_t}\mbx_0+\sqrt{1-\Bar{\alpha}_t} \mbepsilon$, $\mbepsilon_\theta(\mbx_t,t) =-\sqrt{1-\Bar{\alpha}_t} \mbs_\theta(\mbx_t,t)$. With these two approximation, we can learn $q_\theta(\mbx_0)$ to estimate unknown data distribution $p_0(\mbx_0)$ solving \cref{eq: reverse sde} from $t = T$ to $t=0$.

\subsection{Guided Sampling}
\label{sec: prel guided sampling }
In prior diffusion-based controllable generation frameworks~\cite{janner2022planning,zhong2023guided, rempe2023trace, jiang2023motiondiffuser,lin2024joint}, controllable generation is achieved via biasing of the score function for sampling:
\begin{equation}
    \nabla_{\mbx_t} \log \left[p_t(\mbx_t)\exp(-\mathcal{C}(\mbx_t))\right]=\nabla_{\mbx_t} \log p_t(\mbx_t) - \nabla_{\mbx_t}  \mathcal{C}(\mbx_t)
\end{equation}
where $\mathcal{C}(\cdot)$ is the guidance function. It requires estimating the guidance gradient with respect to the noisy data $\mbx_t$. Some approaches introduce an additional neural network to approximate $\mathcal{C}(\cdot)$ at different noisy levels~\cite{dhariwal2021diffusion,janner2022planning,zhong2023guided}. The additional neural network imposes additional training costs and heavier computational burden at the inference stage. To this end, some works define an analytical guidance function $\cJ(\cdot)$ on the clean data $\mbx_0$. They first estimate $\hat{\mbx}_0=f_\theta(\mbx_t)$ based on $\mbx_t$ with the diffusion model, then calculate $\mathcal{C}(\mbx_t)$ as $\cJ(\hat{\mbx}_0)$. However, when taking the gradient of $\cJ(\hat{\mbx}_0)$ with respect to $\mbx_t$, we get 
\begin{equation}
\nabla_{\mbx_t}\cJ(\hat{\mbx}_0)=\nabla_{\hat{\mbx}_0}\cJ(\hat{\mbx}_0) \cdot \nabla_{\mbx_t}f_\theta(\mbx_t).
\end{equation}
Estimating $\nabla_{\mbx_t}f_\theta(\mbx_t)$ requires backpropagating throughout the entire diffusion model, i.e., $f_\theta(\mbx_t)$. It requires heavy computing resources and GPU memory.

\subsection{Trajectory Prediction and Controllable Generation}
Joint trajectory prediction aims to predict the future joint trajectories $\mbx_0$ for all the vehicles in the scene, conditioned on context information $\mbc$. It can be regarded as a conditional generation task where the goal is to train a generative model to approximate the distribution $p_0(\mbx_0|\mbc)$. For simplicity, we omit $\mbc$ and represent $p_0(\mbx_0|\mbc)$ as $p_0(\mbx_0)$. We denote $n$ as the number of vehicles in the same scene and $\mbx_{0,i}$ as the future trajectory for vehicle $i$, $i\in \{1,2,...,n\}$, so $\mbx_0=[\mbx_{0,1},\mbx_{0,2},...,\mbx_{0,n}]$. Joint trajectory prediction can be very challenging. The complex interactions among vehicles, especially in highly interactive and dense traffic, result in a complicated high-dimensional $p_0(\mbx_0)$, which is difficult to accurately model with lightweight and computationally efficient models. A simplified solution is to approximately decompose the joint trajectory distribution into marginal ones, i.e., $p_0(\mbx_{0})\approx \prod_i^n p_0(\mbx_{0,i})$. It leads to the marginal trajectory prediction task, which has been extensively studied with mature solutions~\cite{nayakanti2023wayformer, zhou2023query, lan2023sept}. One drawback is that it omits the interactions among vehicles in the predicted horizon, which leads to large errors in highly interactive scenes.

Controllable trajectory generation is closely related to trajectory prediction. In addition to generating realistic trajectory samples resembling the ground-truth $p_0(\mbx_0)$, controllable generation imposes an additional objective\textemdash the generated trajectories should comply with specified guidance cost function $\cJ(\cdot)$. We term the former objective as \textit{realism} and the latter as \textit{guidance effectiveness}. The guidance cost function $\cJ(\cdot)$ can be some goal points of the vehicles, kinematic constraints, etc, which are mostly defined on the clean manifold rather than on the noisy data when it comes to diffusion-based controllable generation.

\section{Methodology}
\label{sec:method}
\subsection{Optimal Gaussian Diffusion}

As discussed in \cref{sec: pre diffusion}, diffusion models typically select a non-informative prior distribution $p_{prior}$, such as a standard Gaussian, as the initial data distribution for the reverse diffusion process. Such a non-informative $p_{prior}$ is reasonable since $p_T(\mbx_T)$ converges to it when $T \rightarrow \infty$. However, it also means that a large $T$ is required at inference time to yield good performance, which undermines computational efficiency and limits its wide real-time applications in autonomous driving. In this section, we aim to investigate a practical solution to tackle this challenge for joint trajectory prediction. Given the inherent limitation imposed by a non-informative prior, the key question we look into is: \emph{can we 
instead adopt an informative prior so that we can obtain the same level of performance with much smaller reverse diffusion steps?}

First, we still consider a Gaussian prior, but with learnable parameters, i.e., we parameterize $p_
{prior}$ as $q_\phi(T)=\mathcal{N}(\mbmu,\mbSigma)$, where $\mbmu$ and $\mbSigma$ are learnable. We aim to optimize $\mbmu$ and $\mbSigma$ to enhance the generation performance for small $T$. We got inspiration from~\cite{franzese2023much}, where an upper bound of the Kullback–Leibler divergence of the clean data distribution $p_0(\mbx_0)$ and learned distribution $q_\theta(\mbx_0)$ was derived as a function of the diffusion time $T$: 
\begin{equation}
    \text{KL}[p_0(\mbx_0)||q_\theta(\mbx_0)]\leq\mathcal{G}(\mbx_\theta, T)+\text{KL}[p_T(\mbx_T)||p_{prior}]
\end{equation}
where $\mathcal{G}(\mbx_\theta, T)$ is the positive accumulated error between $\nabla_\mbx \log p_t(\mbx_t)$ and $\mbs_\theta(\mbx_t,t)$ from $0$ to $T$ \cite{franzese2023much}. Note that $\mathcal{G}(\mbx_\theta, T)$ is an accumulated error so $\mathcal{G}(\mbx_\theta, T_1) \leq \mathcal{G}(\mbx_\theta, T_2)$, $T_1 \leq T_2$. If we can achieve lower $\text{KL}[p_{T_1}(\mbx_{T_1})||q_\phi({T_1})]$, then a tighter upper bound can be obtained. This opens up the possibility to achieve better performance with less diffusion time. Thus, we propose to optimize the prior distribution by minimizing $\text{KL}[p_T(\mbx_T)||q_\phi(T)]$. As shown in \cref{Proposition: uc opd} (See proof in \cref{app: proof}), it turns out that the optimal $\mu$ and $\Sigma$ can be expressed analytically as functions of the ground-truth data statistics. In addition, we find that we can further minimize the target KL divergence if we set a learnable $\mbSigma_p$ in the perturbation kernel $p_t(\mbx_t|\mbx_0) = \mathcal{N}(\mbx_t;\sqrt{\Bar{\alpha}_t} \mbx_0, (1-\Bar{\alpha}_t)\mbSigma_p)$, whose optimal value can also be expressed as a function of the data statistics. 

\begin{proposition}
\label{Proposition: uc opd} 
Denote $\mbmu_d$ and $\mbSigma_d$ as the mean and variance of $p_{data}$. Denote $\mbSigma^*(i,j)$ and $\mbSigma_p^*(i,j)$ as the element at $i$th row and $j$th column of matrix $\mbSigma^*$ and $\mbSigma_p^*$. The optimal solution to $\min \text{KL}[p_T(\mbx_T)||q_\phi(\mbx_T,T)]$ is 
\begin{equation}
    \begin{cases}
        \mbmu^* \approx \sqrt{\Bar{\alpha}_T} \mbmu_d \\
        \mbSigma_p^*(i,j) \approx \frac{1}{|\mbSigma_d|}\mbSigma_d(i,j)\\
        \mbSigma^*(i,j) \approx \Bar{\alpha}_T\mbSigma_d+(1-\Bar{\alpha}_T)^2\mbSigma_p^* = (\Bar{\alpha}_T^2+\frac{(1-\Bar{\alpha}_T)^2}{|\mbSigma_d|})\mbSigma_d(i,j)\\
    \end{cases}
\end{equation}
\end{proposition}

Thus, if we are able to estimate the mean and variance of the ground-truth data distribution, we can then analytically determine the optimal prior distribution $\mathcal{N}(\mbmu^*,\mbSigma^*)$ and the optimal perturbation kernel variance $\mbSigma^*_p$ at \emph{any noise level} $T \in [0,\infty)$. It leads to a crucial advantage against prior efforts on prior learning~\cite{zheng2022truncated}, where a target noise level has to be determined at training time in order to train an additional neural network to represent the prior distribution at the pre-selected noise level. In contrast, our method enables flexible tuning of the number of diffusion steps at inference time, without additional training costs. Since the learnable $q_\phi(T)$ and perturbation kernel are both Gaussian, we refer to our proposed diffusion framework as Optimal Gaussian Diffusion (OGD).

For joint trajectory prediction and generation, we need to estimate the mean and variance of the joint trajectory distribution $p_0(\mbx_0)$. It is not straightforward as only a limited number of trajectory samples exist in the dataset under the same context. Considering there exists mature and accurate marginal trajectory prediction models \cite{zhou2023query, nayakanti2023wayformer,lan2023sept}, we can conveniently extract statistics of the marginal trajectory distributions, i.e., $p_0(\mbx_{0, i}), i\in \{1,2,...,n\}$ from a pre-trained marginal trajectory predictor. Since \cref{Proposition: uc opd} provides element-wise optimal value, we can easily get:

\begin{corollary} 
\label{Corollary: ogd marginal only} 
Denote $\mbmu_d(\mbx_{0,i})$ and $\mbSigma_d(\mbx_{0,i})$ as the mean and variance of marginal distribution for vehicle $i$ and set both $\mbSigma$ and $\mbSigma_p$ are block-diagonal matrices where each block represents the marginal characteristics of one single vehicle. The optimal solution to $\min \text{KL}[p_T(\mbx_T)||q_\phi(\mbx_T,T)]$ is

\begin{equation}
    \begin{cases}
        \mbmu^*=[\sqrt{\Bar{\alpha}_T}\mbmu_d(\mbx_{0,1}),...,\sqrt{\Bar{\alpha}_T}\mbmu_d(\mbx_{0,n})]^T \\
        \mbSigma_p^* =\frac{1}{\sum_{i=1}^n|\mbSigma_d(\mbx_{0,i})|} \text{diag}[\mbSigma_d(\mbx_{0,1}),...,\mbSigma_d(\mbx_{0,n})]\\
        \mbSigma^*=\Bar{\alpha}_T\text{diag}[\mbSigma_d(\mbx_{0,1}),...,\mbSigma_d(\mbx_{0,n})] + (1-\Bar{\alpha}_T)^2 \mbSigma_p^*\\
    \end{cases}
\end{equation}
\end{corollary}

\cref{Corollary: ogd marginal only} implies that, if we further confine $\mbSigma$ and $\mbSigma_p$ to be block-diagonal without covariance between the states of different vehicles, then we can determine their optimal values purely from the estimated marginal statistics $\mbmu_d(\mbx_{0,i})$ and $\mbSigma_d(\mbx_{0,i})$, $i\in\{1,2,...,n\}$, which enables a practical implementation of the proposed OGD model for joint trajectory prediction and generation tasks. Specifically, for vehicle $i$, we leverage a pre-trained marginal trajectory predictor \cite{zhou2023query, nayakanti2023wayformer,lan2023sept}, predict diverse marginal trajectory sample set $\mathcal{R}_i=\{\mbr_i^l\}_{l=1}^L$ and corresponding likelihood set $\{p(\mbr_i^l)\}_{l=1}^L$, and estimate $\mbmu_d(\mbx_{0,i})$ and $\mbSigma_d(\mbx_{0,i})$. For example, $\mbmu_d(\mbx_{0,i})$ can be estimated as $\frac{1}{L}\sum_{l=L}p(\mbr_i^l)\mbr_i^l$. 

\subsection{Estimated Clean Manifold Guidance with Reference}
As discussed in \cref{sec: prel guided sampling }, the intensive computation required for guided sampling comes from the calculation of $\nabla_{\mbx_t}f_\theta(\mbx_t)$. Previous guided sampling approaches bias the score function defined on the intermediate noisy data $\mbx_t$. In this section, we aim to investigate whether we can inject the gradient directly into $\mbx_0$ to avoid the gradient propagation process. 
We first reformulate the controllable generation as a multi-objective optimization problem directly over $\mbx_0$ and propose an iterative algorithm to solve the formulated problem. In addition, we use reference trajectory points to create the region of interest, which helps solve the local optimal problem caused by multi-modal joint trajectory distribution and accelerate the guided sampling process.

\subsubsection{Estimated Clean Manifold Guidance.}
\label{sec: ecm}
The objective of controllable generation of sample $\mbx_0$ includes two different objectives. The most important is the negative likelihood, ensuring the sample lies in the clean manifold. The second important is the guidance cost representing the user preference on the generated sample $\mbx_0$. This multi-objective optimization problem can be represented as
\begin{equation}
    \label{eq: problem}
    \min_{\mbx_0} \quad [-\log q_\theta(\mbx_0),\cJ(\mbx_0)]^T
\end{equation}
Inspired by lexicographic optimization \cite{sherali1983preemptive}, we solve this multi-objective optimization problem hierarchically. The main idea is to optimize each objective in the order of importance regardless of the degradation of the other less significant objectives. We first optimize the most important objective, $-\log q_\theta(\mbx_0)$, to generate realistic and high-likelihood samples. The diffusion model achieves this goal effectively by reversing the diffusion process from noisy samples at specific noise level. However, exact noise level of current sample $\mbx_0$ is unknown. To address this, we inject noise at level $t$ into $\mbx_0$, and then denoise it from $t$ with learned diffusion model. This approach, similar to \textit{noise injection and denoising} \cite{song2023consistency,meng2022sdeditguidedimagesynthesis, avrahami2022blended}, improves the desired sample quality.
We iteratively repeat this process $K$ times, injecting guidance at each iteration to strengthen user preference. 

Specifically, denote $\mbx_0(k)$ as the sample at iteration $k$. We first regenerate high-likelihood sample $\hat{\mbx}_0(k)$ by diffusion model:
\begin{equation}
    \label{eq: ecm update likelihood}
    \hat{\mbx}_0(k) \leftarrow \mathbb{E}[q_\theta(\mbx_0(k)|\mbx_{t_k}(k))]=\frac{1}{\sqrt{\Bar{\alpha}_{t_k}}}(\mbx_{t_k}-\sqrt{1-\Bar{\alpha}_{t_k}} \mbepsilon_\theta(\mbx_{t_k}(k),{t_k})),
\end{equation}
where $\mbx_{t_k}(k) = \sqrt{\Bar{\alpha}_{t_k}}\mbx_0(k)+\sqrt{1-\Bar{\alpha}_{t_k}}\mbepsilon$, $\mbepsilon \sim \mathcal{N}(\mathbf{0},\mbSigma_p)$. $t_k \in [0,T]$ is a tunable parameter. Then we minimize the guidance cost function with a small degradation of the most important objective $-\log q_\theta(\mbx_0)$, 
\begin{equation}
    \mbx_0(k-1) \leftarrow \hat{\mbx}_0(k)-\zeta\nabla_{\hat{\mbx}_0(k)}\cJ(\hat{\mbx}_0(k))
\end{equation}
where $\zeta$ is the step size. Small degradation of $-\log q_\theta(\mbx_0)$ is realized by one-step gradient update and proper step size $\zeta$. See the derivation of the optimization process and the parameter tuning in \cref{app: ecm}.

During the iterations, $\mbx_0(k)$, $\forall k \in \{0,1,...,K-1\}$ are not exactly on the clean manifold. We minimize its negative log-likelihood through diffusion model, resulting in it lying on an estimated clean manifold. Thus, we call our method Estimated Clean Manifold (ECM) Guidance.

\subsubsection{Reference Joint Trajectories.} 
\label{sec: ref pts}
\begin{figure*}[t]
    \centering
    \includegraphics[width=0.9\linewidth]{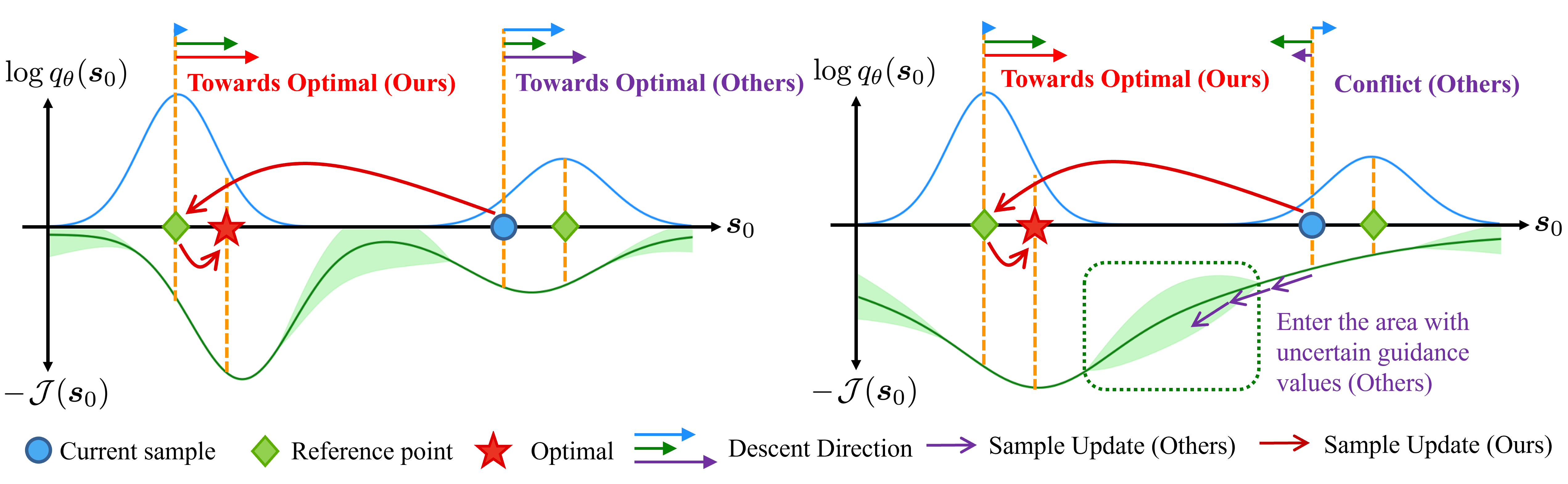}
    \caption{Two challenges with multi-peak function optimization: 1) Gradients may lead to suboptimal local optima (left); 2) There exist regions with low likelihood but high guidance cost uncertainty, leading to instability (right). Our approach can bypass the lengthy paths between peaks, search for better optima, and avoid uncertain areas.}
    \label{fig:reference pts}
\end{figure*}

To generate trajectories with low guidance cost, we are essentially searching low-cost trajectories within the high-likelihood region. At the same time, joint trajectory distribution is a multi-modal distribution resulting from road topologies and different decision variables, meaning the likelihood has multiple peaks. This leads to the optimal solution of Problem \ref{eq: problem} having multiple local optimals, and each optimal is far away from the other. Guided sampling methods \cite{janner2022planning, zhong2023guided, rempe2023trace, jiang2023motiondiffuser}, including our method ECM, suffer from two challenges (See \cref{fig:reference pts}): 1) it can be trapped at the local optimal around the initial position; 2) it takes massive efforts to drag the sample from one peak to another and transferring from one modal to another will need to pass through the region of low-likelihood (off clean-manifold), leading to numerical instability. 

\begin{wrapfigure}[14]{r}{0.6\textwidth}
\begin{algorithm}[H]
        \scriptsize
	\renewcommand{\algorithmicrequire}{\textbf{Input:}}
	\renewcommand{\algorithmicensure}{\textbf{Output:}}
	\caption{ECMR}
	\label{alg: ecmr}
	\begin{algorithmic}[1]
        \REQUIRE $\cJ(\cdot)$, $\{\Bar{\alpha}_t, \beta_t\}_{t=0}^{T-1}$, $\{t_k\}_{k=0}^{K-1}$
        
        \STATE $\mbx_0(K) \sim \mathcal{N}(0,\mbSigma_p)$
        \FOR {$k=K-1,...,1$}
            \STATE $\mbepsilon \sim \mathcal{N}(0,\mbSigma_p)$
            \STATE $\mbx_{t_k}(k) = \sqrt{\Bar{\alpha}_{t_k}}\mbx_0(k)+\sqrt{1-\Bar{\alpha}_{t_k}}\mbepsilon$

            \STATE $\hat{\mbx}_0(k) = \frac{1}{\sqrt{\Bar{\alpha}_{t_k}}}(\mbx_{t_k}(k)-\sqrt{1-\Bar{\alpha}_{t_k}} \mbepsilon_\theta(\mbx_{t_k}(k),t))$

            \STATE $\hat{\mbx}_0(k)=\argmin \cJ(w), w \in \mathcal{R} \bigotimes \hat{\mbx}_0(k)$
            
            \STATE $\mbx_0(k-1) \leftarrow \hat{\mbx}_0(k)-\zeta\nabla_{\hat{\mbx}_0(k)}\cJ(\hat{\mbx}_0(k))$
        \ENDFOR
		\ENSURE $\mbx_0(0)$
	\end{algorithmic}  
\end{algorithm}
\end{wrapfigure}

To overcome this, we generate high-likelihood reference joint trajectories, choose the best one as the initialization. Note that combinations of samples with high marginal likelihood tend to exhibit high joint likelihood. Therefore, we can utilize the marginal sample set $\mathcal{R}=\{\mathcal{R}_i\}_{i=1}^n$ obtained from pre-trained marginal models to generate the references. Specifically, for iteration $k$, we construct candidate joint trajectory set $\mathcal{R} \bigotimes \hat{\mbx}_0(k) = \{[\mbw_1,\mbw_2,...,\mbw_n]|\mbw_i \in \mathcal{R}_i \cup \{\hat{\mbx}_{0,i}\},i=1,2,...,n\}$. We calculate the guidance cost of all possible combinations, $\cJ(w)$, $w \in \mathcal{R} \bigotimes \hat{\mbx}_0(k)$, and choose the minimal-cost one as the reference. 
The guided sampling algorithm, ECM with reference joint trajectories (ECMR), is introduced in \cref{alg: ecmr}. 

\section{Experiments}
\label{sec: exp}

\subsection{Experimental setup}
{\bf Dataset.} We use Argoverse 2 \cite{wilson2023argoverse}, a widely used and large-scale trajectory prediction dataset, to test the effectiveness of our approaches for joint trajectory prediction and controllable generation. It has a large observation window of 5s and a long prediction horizon of 6s.

{\bf Implementation Details.}
We use the fixed scene context encoder of pre-trained QCNet \cite{zhou2023query} to extract compact and representative context features from context information $\mbc$. Then, we utilize a cross-attention layer to update the intermediate noisy data $\mbx_t$ with multiple contexts, including the history encodings of the target agent, the map encodings, the neighboring agents’ encodings. Inspired by \cite{gilles2021thomas}, we also add a cross-attention layer to update $\mbx_t$ with the diverse marginal trajectory samples $\mbr_i^l$ and its corresponding likelihood. In addition, we use self-attention to allow the interaction between $\mbx_{t,i}$ and $\mbx_{t,j}$. Then the model predicts the noise $\mbepsilon_\theta(\mbx_t,t)$. According to \cite{jiang2023motiondiffuser}, compact trajectory representation helps the diffusion model to generate high-quality trajectories efficiently. Inspired by this, we also learn a linear mapping between the 10-dimensional latent and 120-dimensional trajectories. Similar to \cite{jiang2023motiondiffuser}, we design a rapid sample clustering algorithm so that we can generate a representative joint trajectory set. To increase the efficiency of sampling, we use DDIM \cite{song2020denoising} to accelerate the inference, and the DDIM step stride is 10. See \cref{app: additional implementation and ana} for details and analysis.

\subsection{Joint trajectory prediction}
\label{sec: exp jtp}

We now evaluate OGD for joint trajectory prediction. Given $K$ joint trajectories, the evaluation metrics are 1) $\textbf{avgMinFDE}_K$/$\textbf{avgMinADE}_K$: the average of lowest final/average displacement error (FDE/ADE) of joint trajectory samples; 2) $\textbf{actorMR}_K$: the rate of trajectory predictions that are considered to be ``missed'' (>2m FDE) in the lowest minFDE joint trajectory samples; 3) $\textbf{actorCR}_K$: the rate of collisions across ``best'' (lowest avgMinFDE) joint trajectory samples; 4) $\textbf{avgBrierMinFDE}_K$: calculated similarly to $\text{avgMinFDE}_K$ but scaled by the probability score of joint trajectory samples. We denote metrics with superscript ``*'' as those after sample clustering (see \cref{appendix:cluster}). 

\begin{figure}[tb]
\centering
\begin{minipage}{0.38\linewidth}
  \centering
  \includegraphics[width=\linewidth]{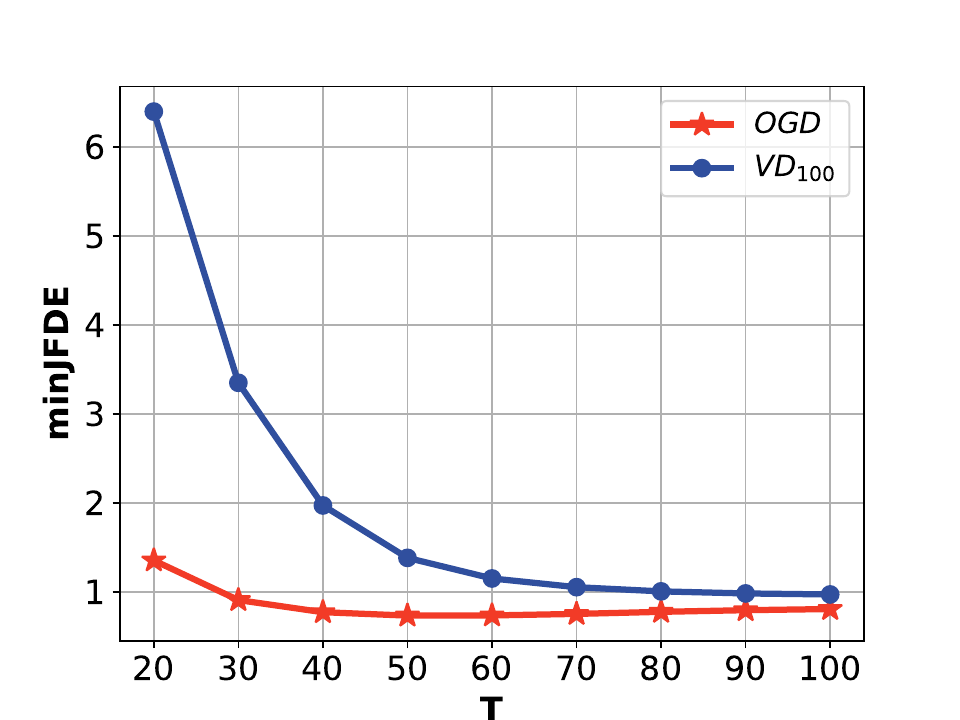}
  \caption{Evaluation of Optimal Gaussian Diffusion and vanilla diffusion over reverse steps $T$.}
  \label{fig: ogd over T}
\end{minipage}
\hfill
\begin{minipage}{0.6\linewidth}
\centering
\tabcaption{\footnotesize Evaluation on joint trajectory prediction task. For each metric, the best result is in \textbf{bold} and the second best result is \underline{underlined}. $T=70$ is the best $T$ from \cref{fig: ogd over T}. $T=40$ is the minimal diffusion time when OGD outperforms $\text{VD}_{500}$ on all metrics.}
\label{tb: ogd vs vd}
\resizebox{\linewidth}{!}{
\begin{tabular}{cccccc}
\toprule
Model&$T$&$\text{avgMinFDE}_{6}^*$&$\text{avgMinADE}_{6}^*$&$\text{avgMinFDE}_{128}$&$\text{avgMinADE}_{128}$\\
\midrule
$\text{VD}_{100}$ &100&0.62&1.38&0.49&0.97\\
$\text{VD}_{500}$ &500&0.61&1.36&0.48&0.91\\
\midrule
OGD&100&\underline{0.60}&\underline{1.32}&\underline{0.43}&0.81\\
OGD&70&\textbf{0.59}&\textbf{1.31}&\textbf{0.42}&\textbf{0.75}\\
OGD&40&0.61&1.34&0.47&\underline{0.77}\\
\bottomrule
\end{tabular} 
}
\end{minipage}
\end{figure}

As a baseline, we train a vanilla diffusion (VD) baseline that shares the same neural network architecture as OGD. Specifically, we train an Optimal Gaussian Diffusion model with diffusion time $T_{train}=100$ (OGD), vanilla diffusion with two different diffusion times $T_{train}=100,500$ ($\text{VD}_{100}$, $\text{VD}_{500}$). Note that $T$ is denoted as the diffusion time from which the reverse diffusion process starts. First, during the inference, we change $T$ and evaluate OGD and $\text{VD}_{100}$ who have the same $T_{train}=100$. \Cref{fig: ogd over T} shows that, with the decrease of the reverse steps, $\text{avgMinFDE}_{128}$ of OGD keeps lower than $\text{VD}_{100}$ and it is more stable with the change of $T$. It is also interesting to note that the performance of OGD even becomes better in the early stage. We hypothesize that it is because that $\text{KL}[p_T(\mbx_T)||p_{prior}]$ of OGD is small and the accumulated error of $\mathcal{G}(\mbx_\theta, T)$ is sufficiently reduced by a small $T$. It also shows an advantage of OGD, which is that it is easy for OGD to tune a suitable $T$ for reverse diffusion without training an additional model for every $T$ \cite{zheng2022truncated}. \Cref{tb: ogd vs vd} shows that OGD outperforms $\text{VD}_{500}$ with only 
 $40$ diffusion steps. 

We also compare our OGD with the other state-of-the-art methods on the Argoverse multi-world leaderboard; our approach OGD ranks $4^{\mathrm{th}}$ on the leaderboard ranked by $\text{avgBrierMinFDE}_K$, which demonstrates the effectiveness of our OGD framework. Note that we only list the entries with publications or technical reports in \cref{tb: mtp leaderboard}. Please refer to the official website for the full leaderboard. 

\begin{table*}[b]
\caption{\footnotesize Quantitative results on the Argoverse 2 Multi-world Forecasting leaderboard. For each metric, the best result is in \textbf{bold} and the second best result is \underline{underlined}.}
\label{tb: mtp leaderboard}
\begin{center} {
\resizebox{1.\textwidth}{!}{
\begin{tabular}{cccccccc}
\toprule
Model&$\text{avgMinFDE}_6^*$&$\text{avgMinFDE}_1^*$ &$\text{actorMR}_6^*$&$\text{avgMinADE}_6^*$&$\text{avgMinADE}_1^*$&$\text{avgBrierMinFDE}_6^*$&$\text{actorCR}_6^*$\\
\midrule
QCXet \cite{zhou2023qcnext}&\textbf{1.02}&\textbf{2.29}&\textbf{0.13}&\textbf{0.50}&\textbf{0.94}&\textbf{1.65}&\textbf{0.01}\\
Gnet\cite{gao2023dynamic}&1.46&3.05&0.19&0.69&1.23&2.12&0.01\\
Forecast-MAE \cite{cheng2023forecast}&1.55&3.33&0.19&0.69&1.30&2.24&0.01\\
FJMP\cite{rowe2023fjmp}&1.89&4.00&0.23&0.81&1.52&2.59&0.01\\
\midrule
OGD (Ours) & \underline{1.31} & \underline{2.71}& \underline{0.17} & \underline{0.60} & \underline{1.08} & \underline{1.95} & \underline{0.01} \\
\bottomrule
\end{tabular}} }
\end{center}
\end{table*}

\subsection{Controllable Generation}
\label{sec: exp controllable generation}
{\bf Tasks.} Future behaviors of vehicles can be effectively can be effectively represented by a set of goal points \cite{zhao2021tnt, gu2021densetnt} such as acceleration, braking, and right or left turn. Generating such diverse modes of joint trajectories is a typical motivation for using controllable generation. Thus, we study the controllable generation task where the goal is to reach diverse goal points at a specific time to fully test our guided sampling method. Considering that goal points should lie in realistic routes, and the trajectories to reach such goal points should also lie in such routes, we generate such target goal points in some realistic routes $\text{Route}(\tau_g)$. The guidance cost function can be expressed as
\begin{equation}
    \cJ(\mbx_0)=\frac{1}{n}||\text{Position}(\mbx_{0},\tau_d)-\text{Route}(\tau_g)||_2^2,
\end{equation}
where $\text{Position}(\mbx_{0},\tau_d)$ is the positions of $\mbx_{0}$ at time $\tau_d$. We choose ground-truth trajectories and a random combination set of diverse marginal samples, i.e., $\mathcal{U} = \{[\mbu_1,\mbu_2,...,\mbu_n]|\mbu_i \in \mathcal{R}_i,i=1,2,...,n\}$ as the realistic routes. We denote the former as \textbf{GT} set and the latter as \textbf{U} set. The underlying assumption is that diverse samples from a good marginal trajectory predictor are realistic. We design different velocity settings to cover the diverse controllable generation tasks in autonomous driving: first is \textbf{Normal Speed} (\textbf{N}), $\tau_d=\tau_g=6s$; second is \textbf{Acceleration} (\textbf{A}), $\tau_d=5s<\tau_g=6s$; third is \textbf{Deceleration} (\textbf{D}), $\tau_d=6s<\tau_g=5s$.

{\bf Metrics.} We use the following metrics to evaluate controllable generation performance: Joint Route Deviation Error (JRDE), which measures the displacements to the routes to evaluate the realism, and Joint Final Displacement Error (JFDE), which evaluates the guidance effectiveness. We also evaluate from the ``min'' and ``mean'' perspectives: The ``min'' metric considers the best sample's performance, while the ``mean'' metric assesses the ratio of valid samples.

{\bf Baselines.} 
Guided sampling in controllable generation is mainly divided into two approaches: the first is to directly calculate $\nabla_{\mbx_t} \cJ(\mbx_t)$ \cite{zhong2023guided}; the second is to calculate $\nabla_{\mbx_t}\cJ(\hat{\mbx}_0)$ \cite{rempe2023trace, jiang2023motiondiffuser}. We denote the former as Next Noisy Mean Guidance (NNM) and the latter as Score Function Guidance (SF). For a fair comparison, we use one guidance step followed by one DDIM step. We also tune the gradient step size for different guided sampling with Optimal Gaussian Diffusion and vanilla diffusion and report the results with the optimal step size. See \cref{app: guided sampling baseline and step size} for the details of baseline derivation and step size tuning. We evaluate the following experiments with 128 joint trajectory samples.

\begin{figure*}[t]
    \centering
    \includegraphics[width=1.0\linewidth]{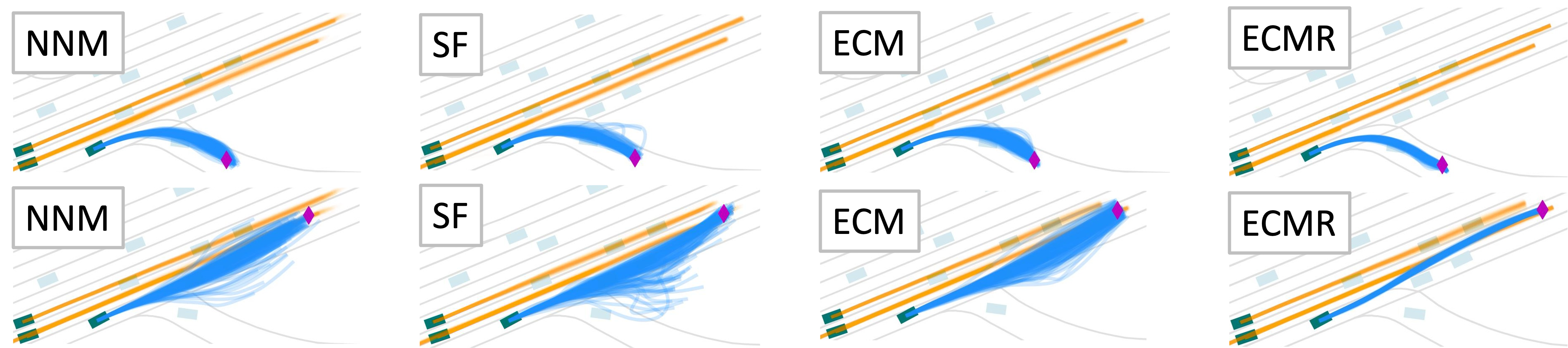}
    \caption{Evaluation on controllable generation: route set \textbf{U} and \textbf{Deceleration}. Magenta diamonds represent goal points. In the first (second) row, goal points are set at the fork lane (right lane). NNM \cite{zhong2023guided} and SF \cite{rempe2023trace, jiang2023motiondiffuser} struggle to drag samples from one modal to another. Our methods can achieve better guidance effectiveness and realism.}
    \label{fig:control generation cases}
\end{figure*}

\begin{table}[tb]
\caption{\footnotesize Evaluation on controllable generation: route set \textbf{U} and \textbf{Deceleration}. For each metric, the best result is in \textbf{bold} and the second best result is \underline{underlined}.}
\label{tb: ogd + ecmr}
\begin{center} {\footnotesize
\begin{tabular}{lcccccccccc}
\toprule
 & \multirow{2}*{Model}& \multirow{2}*{Sampling}& \multirow{2}*{DDIM Steps}&\multicolumn{2}{c}{Guidance Effectiveness}&\multicolumn{2}{c}{Realism}\\
 &\multicolumn{1}{c}{}&\multicolumn{1}{c}{}&\multicolumn{1}{c}{} & \multicolumn{1}{c}{minJFDE} & \multicolumn{1}{c}{meanJFDE} & \multicolumn{1}{c}{minJRDE} & \multicolumn{1}{c}{meanJRDE}\\
\midrule
& $\text{VD}_{500}$&No Guid&50 & 1.961 & 5.229 &0.165&0.492 \\
& $\text{VD}_{500}$&NNM \cite{zhong2023guided}&50 & 0.778 & 2.913 &0.130&0.309 \\
& $\text{VD}_{500}$&SF \cite{rempe2023trace,jiang2023motiondiffuser}&50 & 0.538 & 2.339 &0.158&0.500 \\
\midrule
& OGD&No Guid&10 & 1.772 & 5.172 &0.138&0.469 \\
& OGD& ECM (Ours)& 10& \underline{0.072} & \underline{0.237} &\underline{0.128}&\underline{0.236} \\
& OGD& ECMR (Ours)& 10&\textbf{0.053} & \textbf{0.146} &\textbf{0.110}&\textbf{0.154}\\
\bottomrule
\end{tabular} }
\end{center}
\end{table}

\begin{figure}[tb]
\centering
\begin{minipage}{0.45\linewidth}
  \centering
  \includegraphics[width=1\linewidth]{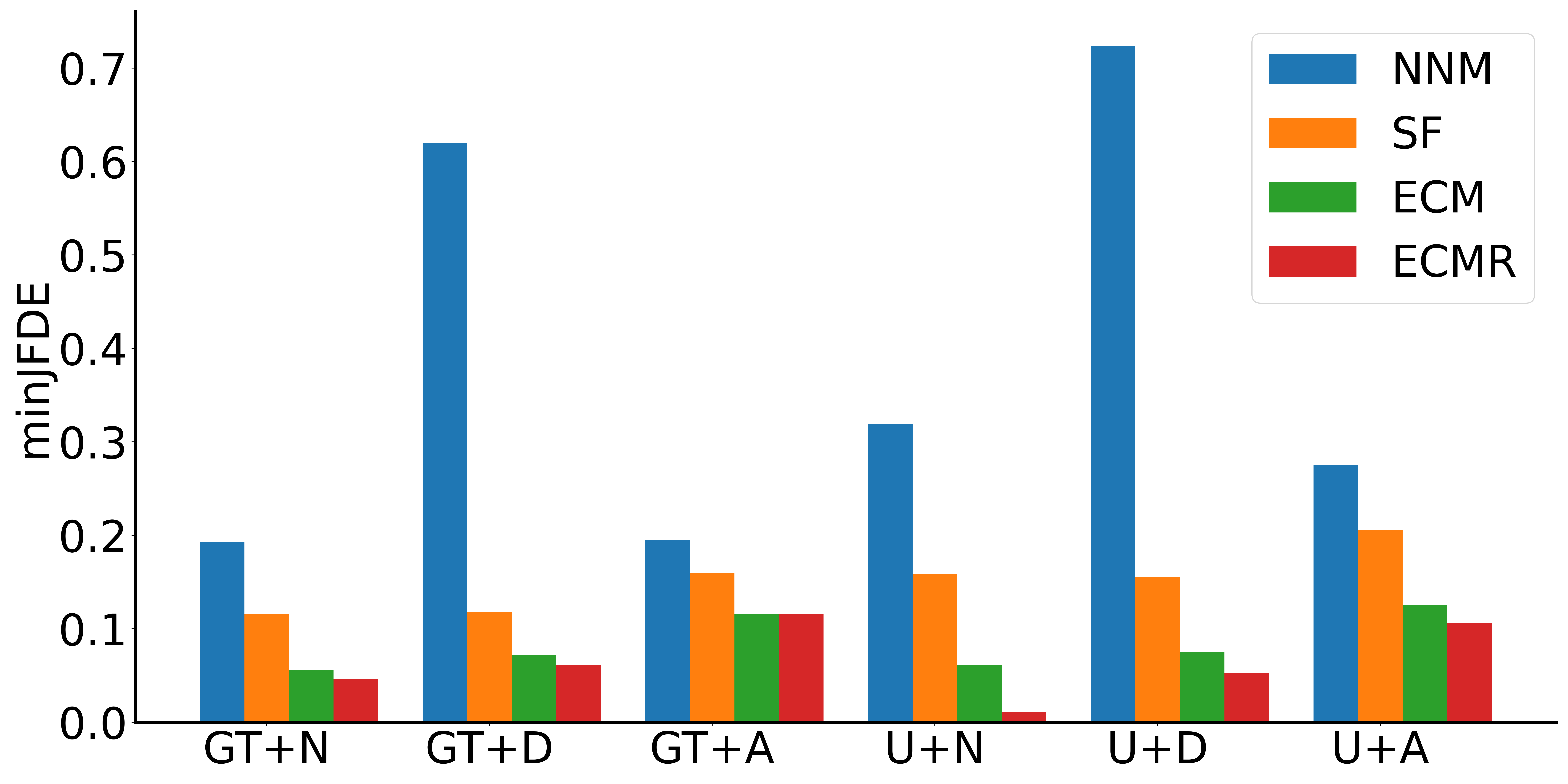}
  \caption{Evaluation of different guided sampling methods on various tasks. See \cref{app: eval on controllable generation task } for other metrics.}
  \label{fig: guided sampling on various tasks }
\end{minipage}
\hfill
\begin{minipage}{0.5\linewidth}
\centering
\tabcaption{\footnotesize We evaluate the average inference time per step and GPU incremental memory on \textbf{U} + \textbf{Deceleration}. We test on single RTX A6000, batch size is 16 and number of samples is 128.}
\label{tb: gpu}
\resizebox{\linewidth}{!}{
\begin{tabular}{cccccc}
\toprule
Sampling&minJFDE&minJRDE&time(ms) &memory (GB)\\
\midrule
NNM \cite{zhong2023guided} &0.724&0.119&113&3.21\\
SF \cite{rempe2023trace, jiang2023motiondiffuser}&0.155&0.126&247&7.96\\
\midrule
ECM(Ours)&0.075&0.116&111&3.21\\
ECMR(Ours)&0.053&0.110&116&3.22\\
\bottomrule
\end{tabular} 
}
\end{minipage}
\end{figure}

{\bf Evaluation.} First, we evaluate the performance and efficiency with our diffusion model (OGD) and guided sampling method (ECM and ECMR) in \cref{tb: ogd + ecmr}, which demonstrates our methods can generate more realistic and effective samples with 5 times less DDIM steps. In addition, with reference joint trajectories, ECM significantly improves 'mean' metrics, indicating it addresses the issues discussed in \cref{sec: ref pts} to a certain extent. Second, we compare solely on different guided sampling methods using the same OGD model shown in \cref{fig: guided sampling on various tasks }. Our ECM achieves better performance both in guidance effectiveness and realism. In \cref{fig:control generation cases}, our ECM can generate trajectories that reach goal points more closely than NNM and SF. And with reference joint trajectory, ECMR can easily move samples from one modal to another . Third, we evaluate the average inference time and average GPU memory usage in \cref{tb: gpu}. Our methods can generate realistic trajectories satisfying guidance quite well with low inference time and GPU memory usage. More results on controllable generation can be found in \cref{app: eval on controllable generation task }.

\section{Conclusion}
In this work, we introduce Optimal Gaussian Diffusion (OGD) and Estimated Clean Manifold (ECM) Guidance to significantly improve the computational efficiency and performance of diffusion models in autonomous driving. These methodologies enable a substantial reduction in inference steps and computational demands while ensuring enhanced joint trajectory prediction and controllable generation capabilities. Our approaches and experimental results underscore the potential of diffusion models for real-time applications in dynamic environments, marking a pivotal advancement in the deployment of diffusion models for autonomous driving. One limitation of the current implementation is that the performance is affected by the accuracy of marginal trajectory predictors. Enhancements could be achieved with superior marginal models or by directly learning joint predictions' mean and variance, areas for future work.

%
%

\section*{Acknowledgement}
This work was supported by Berkeley DeepDrive.
\footnote{https://deepdrive.berkeley.edu}

\bibliographystyle{splncs04}
\bibliography{egbib}

\begin{thebibliography}{10}
\providecommand{\url}[1]{\texttt{#1}}
\providecommand{\urlprefix}{URL }
\providecommand{\doi}[1]{https://doi.org/#1}

\bibitem{avrahami2022blended}
Avrahami, O., Lischinski, D., Fried, O.: Blended diffusion for text-driven editing of natural images. In: Proceedings of the IEEE/CVF conference on computer vision and pattern recognition. pp. 18208--18218 (2022)

\bibitem{cheng2023forecast}
Cheng, J., Mei, X., Liu, M.: Forecast-mae: Self-supervised pre-training for motion forecasting with masked autoencoders. In: Proceedings of the IEEE/CVF International Conference on Computer Vision. pp. 8679--8689 (2023)

\bibitem{chung2022come}
Chung, H., Sim, B., Ye, J.C.: Come-closer-diffuse-faster: Accelerating conditional diffusion models for inverse problems through stochastic contraction. In: Proceedings of the IEEE/CVF Conference on Computer Vision and Pattern Recognition. pp. 12413--12422 (2022)

\bibitem{dhariwal2021diffusion}
Dhariwal, P., Nichol, A.: Diffusion models beat gans on image synthesis. Advances in neural information processing systems  \textbf{34},  8780--8794 (2021)

\bibitem{franzese2023much}
Franzese, G., Rossi, S., Yang, L., Finamore, A., Rossi, D., Filippone, M., Michiardi, P.: How much is enough? a study on diffusion times in score-based generative models. Entropy  \textbf{25}(4), ~633 (2023)

\bibitem{gao2023dynamic}
Gao, X., Jia, X., Li, Y., Xiong, H.: Dynamic scenario representation learning for motion forecasting with heterogeneous graph convolutional recurrent networks. IEEE Robotics and Automation Letters  \textbf{8}(5),  2946--2953 (2023)

\bibitem{gilles2021thomas}
Gilles, T., Sabatini, S., Tsishkou, D., Stanciulescu, B., Moutarde, F.: Thomas: Trajectory heatmap output with learned multi-agent sampling. arXiv preprint arXiv:2110.06607  (2021)

\bibitem{goodfellow2014generative}
Goodfellow, I., Pouget-Abadie, J., Mirza, M., Xu, B., Warde-Farley, D., Ozair, S., Courville, A., Bengio, Y.: Generative adversarial nets. In: Advances in neural information processing systems. pp. 2672--2680 (2014)

\bibitem{gu2021densetnt}
Gu, J., Sun, C., Zhao, H.: Densetnt: End-to-end trajectory prediction from dense goal sets. In: Proceedings of the IEEE/CVF International Conference on Computer Vision. pp. 15303--15312 (2021)

\bibitem{gu2022stochastic}
Gu, T., Chen, G., Li, J., Lin, C., Rao, Y., Zhou, J., Lu, J.: Stochastic trajectory prediction via motion indeterminacy diffusion. In: Proceedings of the IEEE/CVF Conference on Computer Vision and Pattern Recognition. pp. 17113--17122 (2022)

\bibitem{guo2023scenedm}
Guo, Z., Gao, X., Zhou, J., Cai, X., Shi, B.: Scenedm: Scene-level multi-agent trajectory generation with consistent diffusion models. arXiv preprint arXiv:2311.15736  (2023)

\bibitem{ho2020denoising}
Ho, J., Jain, A., Abbeel, P.: Denoising diffusion probabilistic models. Advances in neural information processing systems  \textbf{33},  6840--6851 (2020)

\bibitem{ho2022video}
Ho, J., Salimans, T., Gritsenko, A., Chan, W., Norouzi, M., Fleet, D.J.: Video diffusion models (2022)

\bibitem{hyvarinen2005estimation}
Hyv{\"a}rinen, A., Dayan, P.: Estimation of non-normalized statistical models by score matching. Journal of Machine Learning Research  \textbf{6}(4) (2005)

\bibitem{janner2022planning}
Janner, M., Du, Y., Tenenbaum, J.B., Levine, S.: Planning with diffusion for flexible behavior synthesis. arXiv preprint arXiv:2205.09991  (2022)

\bibitem{jiang2023motiondiffuser}
Jiang, C., Cornman, A., Park, C., Sapp, B., Zhou, Y., Anguelov, D., et~al.: Motiondiffuser: Controllable multi-agent motion prediction using diffusion. In: Proceedings of the IEEE/CVF Conference on Computer Vision and Pattern Recognition. pp. 9644--9653 (2023)

\bibitem{kingma2021variational}
Kingma, D., Salimans, T., Poole, B., Ho, J.: Variational diffusion models. Advances in neural information processing systems  \textbf{34},  21696--21707 (2021)

\bibitem{lan2023sept}
Lan, Z., Jiang, Y., Mu, Y., Chen, C., Li, S.E., Zhao, H., Li, K.: Sept: Towards efficient scene representation learning for motion prediction. arXiv preprint arXiv:2309.15289  (2023)

\bibitem{laumanns2006efficient}
Laumanns, M., Thiele, L., Zitzler, E.: An efficient, adaptive parameter variation scheme for metaheuristics based on the epsilon-constraint method. European Journal of Operational Research  \textbf{169}(3),  932--942 (2006)

\bibitem{lin2024joint}
Lin, H., Wang, Y., Huo, M., Peng, C., Liu, Z., Tomizuka, M.: Joint pedestrian trajectory prediction through posterior sampling. arXiv preprint arXiv:2404.00237  (2024)

\bibitem{lu2022dpm}
Lu, C., Zhou, Y., Bao, F., Chen, J., Li, C., Zhu, J.: Dpm-solver: A fast ode solver for diffusion probabilistic model sampling in around 10 steps. Advances in Neural Information Processing Systems  \textbf{35},  5775--5787 (2022)

\bibitem{luo2021diffusion}
Luo, S., Hu, W.: Diffusion probabilistic models for 3d point cloud generation. In: Proceedings of the IEEE/CVF Conference on Computer Vision and Pattern Recognition. pp. 2837--2845 (2021)

\bibitem{meng2022sdeditguidedimagesynthesis}
Meng, C., He, Y., Song, Y., Song, J., Wu, J., Zhu, J.Y., Ermon, S.: Sdedit: Guided image synthesis and editing with stochastic differential equations (2022), \url{https://arxiv.org/abs/2108.01073}

\bibitem{nayakanti2023wayformer}
Nayakanti, N., Al-Rfou, R., Zhou, A., Goel, K., Refaat, K.S., Sapp, B.: Wayformer: Motion forecasting via simple \& efficient attention networks. In: 2023 IEEE International Conference on Robotics and Automation (ICRA). pp. 2980--2987. IEEE (2023)

\bibitem{ngiam2021scene}
Ngiam, J., Caine, B., Vasudevan, V., Zhang, Z., Chiang, H.T.L., Ling, J., Roelofs, R., Bewley, A., Liu, C., Venugopal, A., et~al.: Scene transformer: A unified architecture for predicting multiple agent trajectories. arXiv preprint arXiv:2106.08417  (2021)

\bibitem{peng2023delflow}
Peng, C., Wang, G., Lo, X.W., Wu, X., Xu, C., Tomizuka, M., Zhan, W., Wang, H.: Delflow: Dense efficient learning of scene flow for large-scale point clouds. In: Proceedings of the IEEE/CVF International Conference on Computer Vision. pp. 16901--16910 (2023)

\bibitem{peng2024q}
Peng, C., Xu, C., Wang, Y., Ding, M., Yang, H., Tomizuka, M., Keutzer, K., Pavone, M., Zhan, W.: Q-slam: Quadric representations for monocular slam. arXiv preprint arXiv:2403.08125  (2024)

\bibitem{rempe2023trace}
Rempe, D., Luo, Z., Bin~Peng, X., Yuan, Y., Kitani, K., Kreis, K., Fidler, S., Litany, O.: Trace and pace: Controllable pedestrian animation via guided trajectory diffusion. In: Proceedings of the IEEE/CVF Conference on Computer Vision and Pattern Recognition. pp. 13756--13766 (2023)

\bibitem{rombach2022high}
Rombach, R., Blattmann, A., Lorenz, D., Esser, P., Ommer, B.: High-resolution image synthesis with latent diffusion models. In: Proceedings of the IEEE/CVF conference on computer vision and pattern recognition. pp. 10684--10695 (2022)

\bibitem{rowe2023fjmp}
Rowe, L., Ethier, M., Dykhne, E.H., Czarnecki, K.: Fjmp: Factorized joint multi-agent motion prediction over learned directed acyclic interaction graphs. In: Proceedings of the IEEE/CVF Conference on Computer Vision and Pattern Recognition. pp. 13745--13755 (2023)

\bibitem{salimans2022progressive}
Salimans, T., Ho, J.: Progressive distillation for fast sampling of diffusion models. arXiv preprint arXiv:2202.00512  (2022)

\bibitem{san2021noise}
San-Roman, R., Nachmani, E., Wolf, L.: Noise estimation for generative diffusion models. arXiv preprint arXiv:2104.02600  (2021)

\bibitem{sherali1983preemptive}
Sherali, H.D., Soyster, A.L.: Preemptive and nonpreemptive multi-objective programming: Relationship and counterexamples. Journal of Optimization Theory and Applications  \textbf{39},  173--186 (1983)

\bibitem{shi2023mtr++}
Shi, S., Jiang, L., Dai, D., Schiele, B.: Mtr++: Multi-agent motion prediction with symmetric scene modeling and guided intention querying. arXiv preprint arXiv:2306.17770  (2023)

\bibitem{song2020denoising}
Song, J., Meng, C., Ermon, S.: Denoising diffusion implicit models. arXiv preprint arXiv:2010.02502  (2020)

\bibitem{song2022pseudoinverse}
Song, J., Vahdat, A., Mardani, M., Kautz, J.: Pseudoinverse-guided diffusion models for inverse problems. In: International Conference on Learning Representations (2022)

\bibitem{song2023consistency}
Song, Y., Dhariwal, P., Chen, M., Sutskever, I.: Consistency models. arXiv preprint arXiv:2303.01469  (2023)

\bibitem{song2019generative}
Song, Y., Ermon, S.: Generative modeling by estimating gradients of the data distribution. Advances in neural information processing systems  \textbf{32} (2019)

\bibitem{song2020score}
Song, Y., Sohl-Dickstein, J., Kingma, D.P., Kumar, A., Ermon, S., Poole, B.: Score-based generative modeling through stochastic differential equations. arXiv preprint arXiv:2011.13456  (2020)

\bibitem{sun2022m2i}
Sun, Q., Huang, X., Gu, J., Williams, B.C., Zhao, H.: M2i: From factored marginal trajectory prediction to interactive prediction. In: Proceedings of the IEEE/CVF Conference on Computer Vision and Pattern Recognition. pp. 6543--6552 (2022)

\bibitem{suo2021trafficsim}
Suo, S., Regalado, S., Casas, S., Urtasun, R.: Trafficsim: Learning to simulate realistic multi-agent behaviors. In: Proceedings of the IEEE/CVF Conference on Computer Vision and Pattern Recognition. pp. 10400--10409 (2021)

\bibitem{sun2022pseudolabel}
Sur, L., Tang, C., Niu, Y., Sachdeva, E., Choi, C., Misu, T., Tomizuka, M., Zhan, W.: Domain knowledge driven pseudo labels for interpretable goal-conditioned interactive trajectory prediction. In: 2022 IEEE/RSJ International Conference on Intelligent Robots and Systems (IROS). pp. 13034--13041 (2022)

\bibitem{varadarajan2022multipath++}
Varadarajan, B., Hefny, A., Srivastava, A., Refaat, K.S., Nayakanti, N., Cornman, A., Chen, K., Douillard, B., Lam, C.P., Anguelov, D., et~al.: Multipath++: Efficient information fusion and trajectory aggregation for behavior prediction. In: 2022 International Conference on Robotics and Automation (ICRA). pp. 7814--7821. IEEE (2022)

\bibitem{watson2022learning}
Watson, D., Chan, W., Ho, J., Norouzi, M.: Learning fast samplers for diffusion models by differentiating through sample quality. arXiv preprint arXiv:2202.05830  (2022)

\bibitem{wilson2023argoverse}
Wilson, B., Qi, W., Agarwal, T., Lambert, J., Singh, J., Khandelwal, S., Pan, B., Kumar, R., Hartnett, A., Pontes, J.K., et~al.: Argoverse 2: Next generation datasets for self-driving perception and forecasting. arXiv preprint arXiv:2301.00493  (2023)

\bibitem{xu2022pretram}
Xu, C., Li, T., Tang, C., Sun, L., Keutzer, K., Tomizuka, M., Fathi, A., Zhan, W.: Pretram: Self-supervised pre-training via connecting trajectory and map. In: European Conference on Computer Vision. pp. 34--50. Springer (2022)

\bibitem{xu2023bits}
Xu, D., Chen, Y., Ivanovic, B., Pavone, M.: Bits: Bi-level imitation for traffic simulation. In: 2023 IEEE International Conference on Robotics and Automation (ICRA). pp. 2929--2936. IEEE (2023)

\bibitem{yin2021diverseinteraction}
Yin, Z.H., Sun, L., Sun, L., Tomizuka, M., Zhan, W.: Diverse critical interaction generation for planning and planner evaluation. In: 2021 IEEE/RSJ International Conference on Intelligent Robots and Systems (IROS). pp. 7036--7043 (2021)

\bibitem{zhang2022fast}
Zhang, Q., Chen, Y.: Fast sampling of diffusion models with exponential integrator. arXiv preprint arXiv:2204.13902  (2022)

\bibitem{zhao2021tnt}
Zhao, H., Gao, J., Lan, T., Sun, C., Sapp, B., Varadarajan, B., Shen, Y., Shen, Y., Chai, Y., Schmid, C., et~al.: Tnt: Target-driven trajectory prediction. In: Conference on Robot Learning. pp. 895--904. PMLR (2021)

\bibitem{zheng2022truncated}
Zheng, H., He, P., Chen, W., Zhou, M.: Truncated diffusion probabilistic models. arXiv preprint arXiv:2202.09671  (2022)

\bibitem{zhong2023guided}
Zhong, Z., Rempe, D., Xu, D., Chen, Y., Veer, S., Che, T., Ray, B., Pavone, M.: Guided conditional diffusion for controllable traffic simulation. In: 2023 IEEE International Conference on Robotics and Automation (ICRA). pp. 3560--3566. IEEE (2023)

\bibitem{zhou2023query}
Zhou, Z., Wang, J., Li, Y.H., Huang, Y.K.: Query-centric trajectory prediction. In: Proceedings of the IEEE/CVF Conference on Computer Vision and Pattern Recognition. pp. 17863--17873 (2023)

\bibitem{zhou2023qcnext}
Zhou, Z., Wen, Z., Wang, J., Li, Y.H., Huang, Y.K.: Qcnext: A next-generation framework for joint multi-agent trajectory prediction (2023)

\end{thebibliography}

\newpage
\appendix
\section{Proof of \cref{Proposition: uc opd}}
\label{app: proof}
In this section, we prove the \cref{Proposition: uc opd}.

\begin{proof}
We would like to minimize $\text{KL}[p_T(\mbx_T)||q_\phi(T)]$ in the VP-SDE ~\cite{song2020score} setting. We expand VP-SDE setting into the setting when the perturbation kernel is in the format of $p_t(\mbx_t|\mbx_0) = \mathcal{N}(\mbx_t;a_t \mbx_0+b_t, c_t^2\mbSigma_p)$. For VP-SDE, $a_t=\sqrt{\Bar{\alpha}_t}$, $b_t=0$, $c_t=1-\Bar{\alpha}_t$. For Variance Exploding (VE) SDE~\cite{song2020score}, $a_t=1$, $b_t=0$, $c_t=\sigma_t^2$. In the following content, we are going to minimize $\text{KL}[p_T(\mbx_T)||q_\phi(T)]$ in a more general form when $\mbx_T = a_T\mbx_0+b_T+\mbepsilon_T$ and $\text{Var}[\mbepsilon_T]=c_T^2\mbSigma_p$. Then, we use the results on VP-SDE.

Minimizing $\text{KL}[p_T(\mbx_T)||q_\phi(T)]$ is equivalent to maximizing the log likelihood $\max_{\mbmu, \mbSigma, \mbSigma_p} \mathbb{E} [\log q_\phi(\mbx_T,T)]$, where $q_\phi(\mbx_T,T)$ is denoted as the probability density of $q_\phi(T) = \mathcal{N}(\mbmu,\mbSigma)$ at $\mbx_T$. Assume the dataset contains $N$ data points, which are $\mbx_0^i, i=1,2,...,N$. And $\mbx_T^i=a_T\mbx_0^i+b_T+\mbepsilon_T^i, i=1,2,...,N$. Thus, $\mathbb{E} [\log q_\phi(\mbx_T,T)]$ can be approximated as $\frac{1}{N}\sum_{i=1}^N \log q_\phi(\mbx_T^i,T)$, denoted as $l(\mbmu, \mbSigma, \mbSigma_p)$. Our optimization problem is formulated as 

\begin{equation}
    \begin{aligned}
        \max_{\mbmu, \mbSigma, \mbSigma_p} & \quad l(\mbmu, \mbSigma, \mbSigma_p)\\
        s.t.& \quad |\mbSigma_p|=1
    \end{aligned}
\end{equation}

We formulate the Lagrange function 
\begin{equation}
    \begin{aligned}
    h(\mbmu, \mbSigma, \mbSigma_p,\lambda) &= l(\mbmu, \mbSigma, \mbSigma_p) + \lambda (\log |\mbSigma_p| - \log 1)\\
    &=l(\mbmu, \mbSigma, \mbSigma_p) + \lambda \log |\mbSigma_p|
    \end{aligned}
\end{equation}

First, we express explicitly $l(\mbmu, \mbSigma, \mbSigma_p)$ as 
\begin{equation}
    \begin{aligned}
        l(\mbmu, \mbSigma, \mbSigma_p)=&\frac{1}{N}\sum_{i=1}^N \log q_\phi(\mbx_T^i,T)\\
        =&-\frac{1}{2} \log |\mbSigma|-\frac{1}{2N}\sum_{i=1}^N(\mbx_T^i-\mbmu)^T\mbSigma^{-1}(\mbx_T^i-\mbmu)\\
        =& -\frac{1}{2} \log |\mbSigma| - \frac{1}{2N}\sum_{i=1}^N(a_T\mbx_0^i+b_T-\mbmu)^T\mbSigma^{-1}(a_T\mbx_0^i+b_T-\mbmu)\\
        &-\frac{1}{N}\sum_{i=1}^N(a_T\mbx_0^i+b_T-\mbmu)^T\mbSigma^{-1} \mbepsilon_T^i-\frac{1}{2N}\sum_{i=1}^N{\mbepsilon_T^i}^T\mbSigma^{-1} \mbepsilon_T^i\\
    \end{aligned}
\end{equation}

Take derivative with respect to $\mbmu$, we get
\begin{equation}
    \begin{aligned}
        \frac{\partial h(\mbmu, \mbSigma, \mbSigma_p,\lambda)}{\partial \mbmu}=&\frac{1}{N}\sum_{i=1}^N(a_T\mbx_0^i+b_T-\mbmu)^T\mbSigma^{-1}+\frac{1}{N}\sum_{i=1}^N{\mbepsilon_T^i}^T\mbSigma^{-1} \\
        =& \frac{1}{N}\sum_{i=1}^N(a_T\mbx_0^i+b_T-\mbmu)^T\mbSigma^{-1} = 0
    \end{aligned}
\end{equation}

In order to take derivative with respect to $\mbSigma$, we reformulate $l(\mbmu, \mbSigma, \mbSigma_p)$ as
\begin{equation}
    \begin{aligned}
        l(\mbmu, \mbSigma, \mbSigma_p)=&\frac{1}{2} \log |\mbSigma^{-1}| - \frac{1}{2N}\sum_{i=1}^N\text{tr}[(a_T\mbx_0^i+b_T-\mbmu)(a_T\mbx_0^i+b_T-\mbmu)^T\mbSigma^{-1}]\\
        &-\frac{1}{N}\sum_{i=1}^N\text{tr}[\mbepsilon_T^i(a_T\mbx_0^i+b_T-\mbmu)^T\mbSigma^{-1}]-\frac{1}{2N}\sum_{i=1}^N\text{tr}[\mbepsilon_T^i{\mbepsilon_T^i}^T\mbSigma^{-1} ]\\
        =&\frac{1}{2} \log |\mbSigma^{-1}| - \frac{1}{2N}\text{tr}[(\sum_{i=1}^N(a_T\mbx_0^i+b_T-\mbmu)(a_T\mbx_0^i+b_T-\mbmu)^T)\mbSigma^{-1}]\\
        &-\frac{1}{N}\text{tr}[(\sum_{i=1}^N\mbepsilon_T^i(a_T\mbx_0^i+b_T-\mbmu)^T)\mbSigma^{-1}]-\frac{1}{2N}\text{tr}[\sum_{i=1}^N(\mbepsilon_T^i{\mbepsilon_T^i}^T)\mbSigma^{-1} ]\\
        \approx & \frac{1}{2} \log |\mbSigma^{-1}| - \frac{1}{2N}\text{tr}[(\sum_{i=1}^N(a_T\mbx_0^i+b_T-\mbmu)(a_T\mbx_0^i+b_T-\mbmu)^T)\mbSigma^{-1}]\\
        &-\text{tr}[\text{Cov}[\mbepsilon, \mbx_T-\mbmu]\mbSigma^{-1}]-\frac{1}{2}\text{tr}[c_T^2\mbSigma_p\mbSigma^{-1} ]\\
        \approx & \frac{1}{2} \log |\mbSigma^{-1}| - \frac{1}{2N}\text{tr}[(\sum_{i=1}^N(a_T\mbx_0^i+b_T-\mbmu)(a_T\mbx_0^i+b_T-\mbmu)^T)\mbSigma^{-1}]\\
        &-\frac{c_T^2}{2}\text{tr}[\mbSigma_p\mbSigma^{-1} ]\\
    \end{aligned}
\end{equation}
where $\text{Cov}[\cdot, \cdot]$ is the covariance matrix of two random variables and $\text{Cov}[\mbx_T-\mbmu, \mbepsilon_T] \approx 0$ because $\mbepsilon_T$ is a independent random variable.

Then take derivative of $\mbSigma^{-1}$, we get:
\begin{equation}
    \begin{aligned}
        \frac{\partial h(\mbmu, \mbSigma, \mbSigma_p,\lambda)}{\partial \mbSigma^{-1}}\approx&\frac{1}{2}\mbSigma-\frac{1}{2N}\sum_{i=1}^N(a_T\mbx_0^i+b_T-\mbmu)(a_T\mbx_0^i+b_T-\mbmu)^T-\frac{c_T^2}{2}\mbSigma_p = 0 
    \end{aligned}
\end{equation}

Take derivative of $\mbSigma_p$
\begin{equation}
    \begin{aligned}
        \frac{\partial h(\mbmu, \mbSigma, \mbSigma_p,\lambda)}{\partial \mbSigma_p} \approx & \frac{\partial[-\frac{c_T^2}{2}\text{tr}[\mbSigma_p\mbSigma^{-1}]+\lambda \log |\mbSigma_p|]}{\partial \mbSigma_p}\\
        =&\frac{\partial[-\frac{c_T^2}{2}\text{tr}[\mbSigma^{-1}\mbSigma_p]+\lambda \log |\mbSigma_p|]}{\partial \mbSigma_p}\\
        =&-\frac{c_T^2}{2} \mbSigma^{-1}+\lambda \mbSigma_p^{-1}=0\\
    \end{aligned}
\end{equation}

Take derivative of $\lambda$
\begin{equation}
    \begin{aligned}
        \frac{\partial h(\mbmu, \mbSigma, \mbSigma_p,\lambda)}{\partial \lambda} = &\log |\mbSigma_p|=0\\
    \end{aligned}
\end{equation}

In this way, we get 
\begin{equation}
    \begin{cases}
        \mbmu^* = b_T + a_T\frac{1}{N}\sum_{i=1}^N\mbx_0^i \\
        \mbSigma^* \approx \frac{1}{N}\sum_{i=1}^N(a_T\mbx_0^i+b_T-\mbmu)(a_T\mbx_0^i+b_T-\mbmu)^T+c_T^2\mbSigma_p^*\\
        \mbSigma_p^* \approx \frac{2\lambda}{c_T^2}\mbSigma^*\\
        |\mbSigma_p^*|=1
    \end{cases}
\end{equation}

Given that $\mbmu_d \approx \frac{1}{N}\sum_{i=1}^N\mbx_0^i$ and $\mbSigma_d \approx \frac{1}{N}\sum_{i=1}^N(\mbx_0^i-\mbmu_d)(\mbx_0^i-\mbmu_d)^T$, we have $\mbSigma^* \approx a_T^2\mbSigma_d+c_T^2\mbSigma_p^*$ and 

\begin{equation}
    \begin{cases}
        \mbmu^* \approx b_T + a_T \mbmu_d \\
        \mbSigma_p^* \approx \frac{1}{|\mbSigma_d|}\mbSigma_d\\
        \mbSigma^* \approx a_T^2\mbSigma_d+c_T^2\mbSigma_p^* = (a_T^2+\frac{c_T^2}{|\mbSigma_d|})\mbSigma_d\\
    \end{cases}
\end{equation}
where the approximation becomes increasingly close to an equality when $N$ is very large.

Denote $\mbSigma^*(i,j)$ and $\mbSigma_p^*(i,j)$ as the element at $i$th row and $j$th column of matrix $\mbSigma^*$ and $\mbSigma_p^*$, and we can have element-wise optimal values for $\mbSigma$ and $\mbSigma_p$,

\begin{equation}
    \begin{cases}
        \mbmu^* \approx b_T + a_T \mbmu_d \\
        \mbSigma_p^*(i,j) \approx \frac{1}{|\mbSigma_d|}\mbSigma_d(i,j)\\
        \mbSigma^*(i,j) \approx a_T^2\mbSigma_d+c_T^2\mbSigma_p^* = (a_T^2+\frac{c_T^2}{|\mbSigma_d|})\mbSigma_d(i,j)\\
    \end{cases}
\end{equation}

The element-wise version is important because perturbation models typically use a diagonal matrix as the diffusion kernel, implying that $\mbSigma_p$ is a diagonal matrix. The element-wise version can handle not only diagonal matrix diffusion kernels but also any other variance matrix diffusion kernels. In a multi-agent setting, if we disentangle the relationships between different agents in the latent space and allow the model to learn these relationships, the variance matrix can be a block diagonal matrix. This means that there are correlations within each agent, but no correlation exists between agents. 

Apply this to VP-SDE and we get \cref{Proposition: uc opd} and \cref{Corollary: ogd marginal only}.
\end{proof}

\section{Derivation of ECM and Parameter Tuning}
\label{app: ecm}
\subsection{Derivation of ECM}
As discussed in \cref{sec: ecm}, we solve the multi-objective problem \ref{eq: problem} by $K$ iterations. In this section, we derive the optimization process for each iteration.

At iteration $k \in \{0,1,...,K-1\}$, we first minimize the most important objective $-\log q_\theta(\mbx_0(k))$ where $\mbx_0(k)$ is the intermediate sample at iteration $k$. Since our ultimate goal in optimizing $-\log q_\theta(\mbx_0(k))$ is to ensure that the sample lies on the clean manifold, we can employ a diffusion model to regenerate the high-likelihood sample based on the current sample $\mbx_0(k)$. We add the noise $\mbepsilon \sim \mathcal{N}(\mathbf{0},\mbSigma_p)$ to the current sample $\mbx_0(k)$ and get $\mbx_{t_k}(k) = \sqrt{\Bar{\alpha}_{t_k}}\mbx_0(k)+\sqrt{1-\Bar{\alpha}_{t_k}}\mbepsilon$ where $t_k \in [0,T]$ is a tunable parameter. In order to obtain $q_\theta(\mbx_0(k)|\mbx_{t_k}(k))$, we can we can iteratively run the reverse diffusion process until $t=0$. However, this approach is time-consuming. Therefore, we use Tweedie's formula to perform a one-step estimation. Specifically, $q_\theta(\mbx_0(k)|\mbx_{t_k}(k))$ can be approximated as a Gaussian distribution and its mean has the highest likelihood. Thus, given $\mbx_{t_k}(k)$, we can update $\mbx_0(k)$ to the highest likelihood position, i.e., $\mathbb{E}[q_\theta(\mbx_0(k)|\mbx_{t_k}(k))]$, 

\begin{equation}
    \label{eq: app ecm update likelihood }
    \hat{\mbx}_0(k) \leftarrow \mathbb{E}[q_\theta(\mbx_0(k)|\mbx_{t_k}(k))]=\frac{1}{\sqrt{\Bar{\alpha}_{t_k}}}(\mbx_{t_k}-\sqrt{1-\Bar{\alpha}_{t_k}} \mbepsilon_\theta(\mbx_{t_k}(k),{t_k}))
\end{equation}

Then, inspired by $\epsilon$-Constraint method \cite{laumanns2006efficient}, we minimize the guidance cost under the constraint of small degradation of the primary objective value,
\begin{equation}
    \label{prob: ecm second}
    \begin{aligned}
        \min_{\mbx_0(k-1)} & \quad \cJ(\mbx_0(k-1))\\
        s.t.& \quad -\log q_\theta(\mbx_0(k-1)) < -\log q_\theta(\hat{\mbx}_0(k))+\gamma\\
    \end{aligned}
\end{equation}
where $\gamma$ is a small positive scalar. We assume that if the deviation of $\mbx_0(k-1)$ from $\hat{\mbx}_0(k)$ is small enough, the constrain in Problem \ref{prob: ecm second} can be satisfied. Thus, we propose a simple approach to solve this problem by one gradient update 
\begin{equation}
    \mbx_0(k-1) \leftarrow \hat{\mbx}_0(k)-\zeta\nabla_{\hat{\mbx}_0(k)}\cJ(\hat{\mbx}_0(k))
\end{equation}
where $\zeta$ is the step size.

\subsection{Parameter tuning and analysis}
In this section, we analyze the impact of $t_k$, where $t_k \in [0,T]$, and injected noise $\mbepsilon \sim \mathcal{N}(\mathbf{0},\mbSigma_p)$ on primary objective: the negative log-likelihood.

First, we examine the impact of $t_k$ where $t_k \in [0,T]$. We first inject noise at level $t_k$ and denoise it from $t_k$ using learned diffusion model. Generally, a larger $t_k$ but highly distinguished samples will negate the guidance efforts made in previous steps. Moreover, a larger $t_k$ reduces sample quality because we use a one-step estimation, and the estimation error increases with a larger $t_k$. Conversely, a smaller $t_k$ introduces minimal noise, facilitating denoising to a high-likelihood sample with only a few denoising steps. Thus, updating with mean of Tweedie’s formula is more accurate. Therefore, we design $t_k$ to decrease as $k$ decrease, enhancing diversity in the initial stage and refining the sample in the final stage.

Second, we analyze the impact of injected noise $\mbepsilon \sim \mathcal{N}(\mathbf{0},\mbSigma_p)$ in $\mbx_{t_k}(k) = \sqrt{\Bar{\alpha}_{t_k}}\mbx_0(k)+\sqrt{1-\Bar{\alpha}_{t_k}}\mbepsilon$. Note that $\mbepsilon$ at different iterations are independent with each other, which injects excessive stochasticity. More importantly, directly injecting stochastic noise at each iteration also diffuse the guidance from previous iterations, resulting in lower guidance effectiveness. Inspired by DDIM \cite{song2020denoising}, we can choose to use $\mbepsilon_\theta(\mbx_{t_{k+1}}(k+1), t_{k+1})$ as the deterministic noise $\mbepsilon$ at iteration $k$. We use Estimated Clean Manifold (ECM) Guidance with the optimal step size tuned in \cref{app: step size tuning}, and compare the performance differences when injecting stochastic versus deterministic noise. The results are shown in \cref{tb: injected noise sto vs det}. We find that ECM with stochastic noise can generate samples as realistic as that with deterministic noise, but the guidance is excessively diffused. We use the deterministic injected noise in this paper.

\begin{table}[tb]
\caption{\footnotesize Comparison of different choices of injected noise. \textbf{GT} denotes the ground truth route set. \textbf{N} signifies normal speed, \textbf{D} indicates deceleration, and 
\textbf{A} stands for acceleration.}
\label{tb: injected noise sto vs det}

\begin{center} {\footnotesize
\begin{tabular}{cccccccc}
\toprule
 & \multirow{2}*{Task}& \multirow{2}*{Noise} &\multicolumn{2}{c}{Guidance Effectiveness}&\multicolumn{2}{c}{Realism}\\
 &\multicolumn{1}{c}{}&\multicolumn{1}{c}{} & \multicolumn{1}{c}{minJFDE} & \multicolumn{1}{c}{meanJFDE} & \multicolumn{1}{c}{minJRDE} & \multicolumn{1}{c}{meanJRDE}\\
\midrule
& \multirow{2}{*}{\textbf{GT}+\textbf{N}}&Stochastic & 0.129 & 0.416 &0.101&0.218 \\
& &Deterministic & 0.056 & 0.209 &0.113&0.232 \\
\midrule
& \multirow{2}{*}{\textbf{GT}+\textbf{D}}&Stochastic & 0.145 & 0.468 &0.124&0.223 \\
& &Deterministic & 0.072 & 0.237 &0.128&0.236 \\
\midrule
& \multirow{2}{*}{\textbf{GT}+\textbf{A}}&Stochastic & 0.444 & 0.979 &0.139&0.218 \\
& &Deterministic & 0.117 & 0.438 &0.137&0.232 \\
\bottomrule
\end{tabular} }
\end{center}
\end{table}

\section{Additional Implementation Details and Analysis}
\label{app: additional implementation and ana}
In this section, we begin by detailing the network structure and parameters such as the diffusion schedule parameters. Subsequently, we describe the process of learning compact trajectory representations through Linear Mapping (LM). Next, we introduce the sample clustering algorithm utilized in this study. Finally, we compare the influence of different pre-trained marginal trajectory predictors.

\subsection{Network Structure and parameters}

We use the fixed scene context encoder of pre-trained QCNet \cite{zhou2023query} to extract compact and representative context features from context information $\mbc$. Then, we utilize a cross-attention layer to update the intermediate noisy data $\mbx_t$ with multiple contexts, including the history encodings of the target agent, the map encodings, the neighboring agents’ encodings. Inspired by \cite{gilles2021thomas}, we also add a cross-attention layer to update $\mbx_t$ with the diverse marginal trajectory samples $\mbr_i^l$ and its corresponding likelihood (generated by pre-trained QCNet \cite{zhou2023query}). In addition, we use self-attention to allow the interaction between $\mbx_{t,i}$ and $\mbx_{t,j}$. Then the model predicts the noise $\mbepsilon_\theta(\mbx_t,t)$.

For the diffusion parameters, we set $\beta_0=0.0001$ and $\beta_T=0.05$ for all the models we trained. And then we calculate $\alpha_t=1-\beta_t$, $\Bar{\alpha}_t=\prod _{t=0}^T\alpha_t$. We train the models for 64 epochs.

For all experiments, we generate 128 samples by default. To facilitate comparison with other state-of-the-art methods on the Argoverse 2 multi-world leaderboard, we generate 2048 joint trajectory samples.

\subsection{Compact Latent Representation}
According to \cite{jiang2023motiondiffuser}, compact and expressive latent representation helps the diffusion model to generate high-quality trajectories efficiently. In this paper, we learn a low-dimensional latent of the trajectory then apply diffusion model in the latent space. For clearer presentation, we use the notation $\mbx$ in this section to represent the trajectory. Denote the latent as $\mbz \in \mathbb{R}^{Z \times 1}$ and the mapping functions between $\mbx$ and $\mbz$ are $\mbx = \mathcal{F}(\mbz)$ and $\mbz=\mathcal{G}(\mbx)$. 

First, we minimize the reconstruction loss, $\mathcal{L}_{rec}$:
\begin{equation}
    \mathcal{L}_{rec}=||\mathcal{F}(\mathcal{G}(\mbx))- \mbx ||_2^2
\end{equation}

Second, we constrain mapping between between $\mbx$ and $\mbz$ should be distance-preserving transformations. The intuition behind is that the latent diffusion model aims to minimize expected error in latent space, though the ultimate objective is reducing error in trajectory space. Consequently, it disproportionately prioritizes samples with significant latent space errors, which may not correspond to those with substantial errors in trajectory space. Thus, we minimize this regularization term $\mathcal{L}_{reg}$:
\begin{equation}
    \mathcal{L}_{reg}=(\mbx^T\mbx - \mathcal{G}(\mbx)^T\mathcal{G}(\mbx))^2
\end{equation}

In addition, extreme variance in different dimension of the latent will cause instability of the training. Thus, we minimize the variance term $\mathcal{L}_{var}$
\begin{equation}
    \mathcal{L}_{var}=||\text{std}(\mathcal{G}(\mbx)) - \mathbf{v}||_2^2
\end{equation}
where $\mathbf{v} \in \mathbb{R}^{Z \times 1}$ whose elements are all $\eta$, a learnable variable.

\begin{figure*}[tb]
    \centering
    \begin{subfigure}{0.29\textwidth}
            \includegraphics[width=\linewidth]{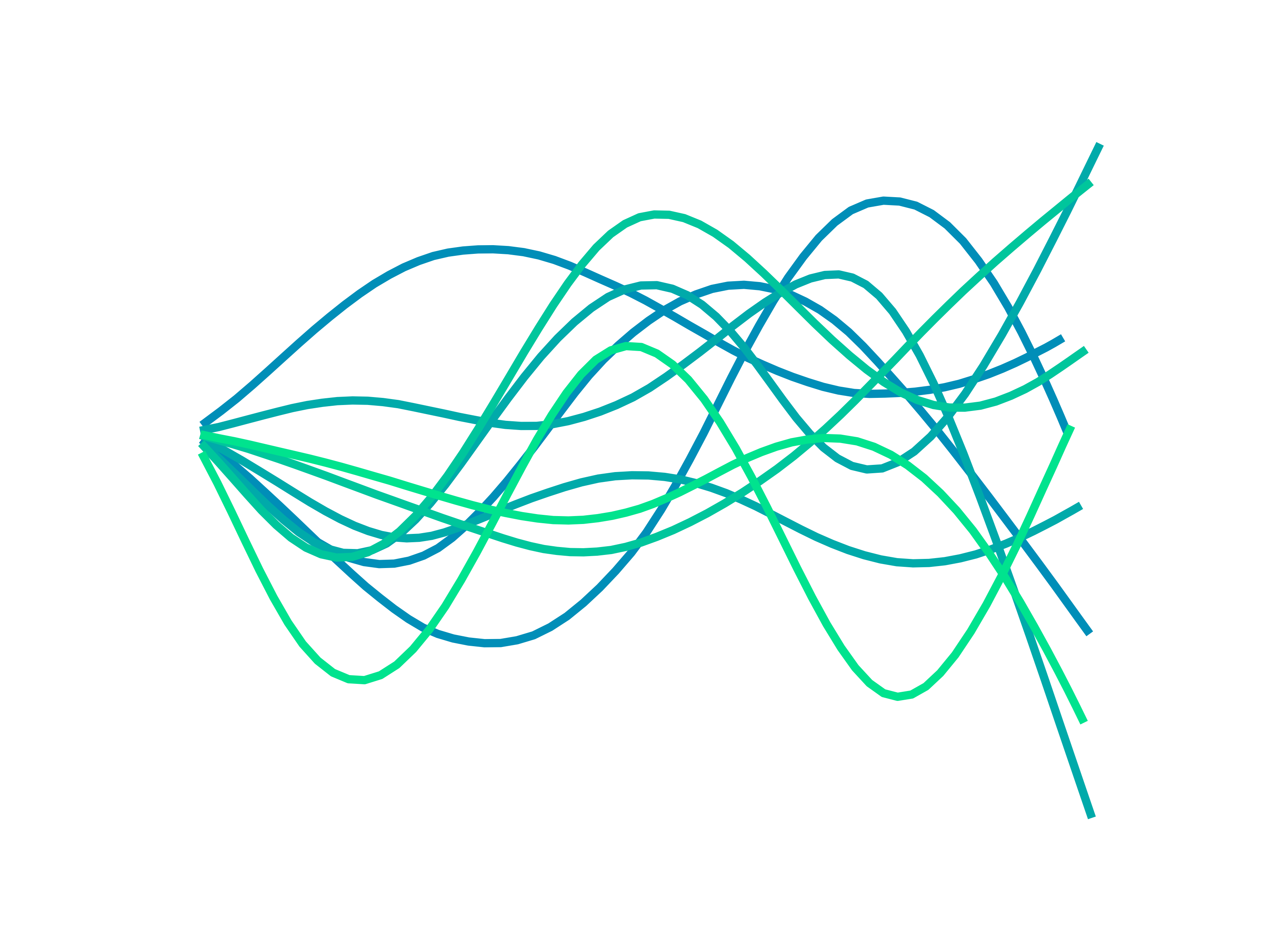}
            \caption{Modes from LM.}
            \label{fig:lm pca}
    \end{subfigure}%
    \hfill
    \begin{subfigure}{0.29\textwidth}
            \includegraphics[width=\linewidth]{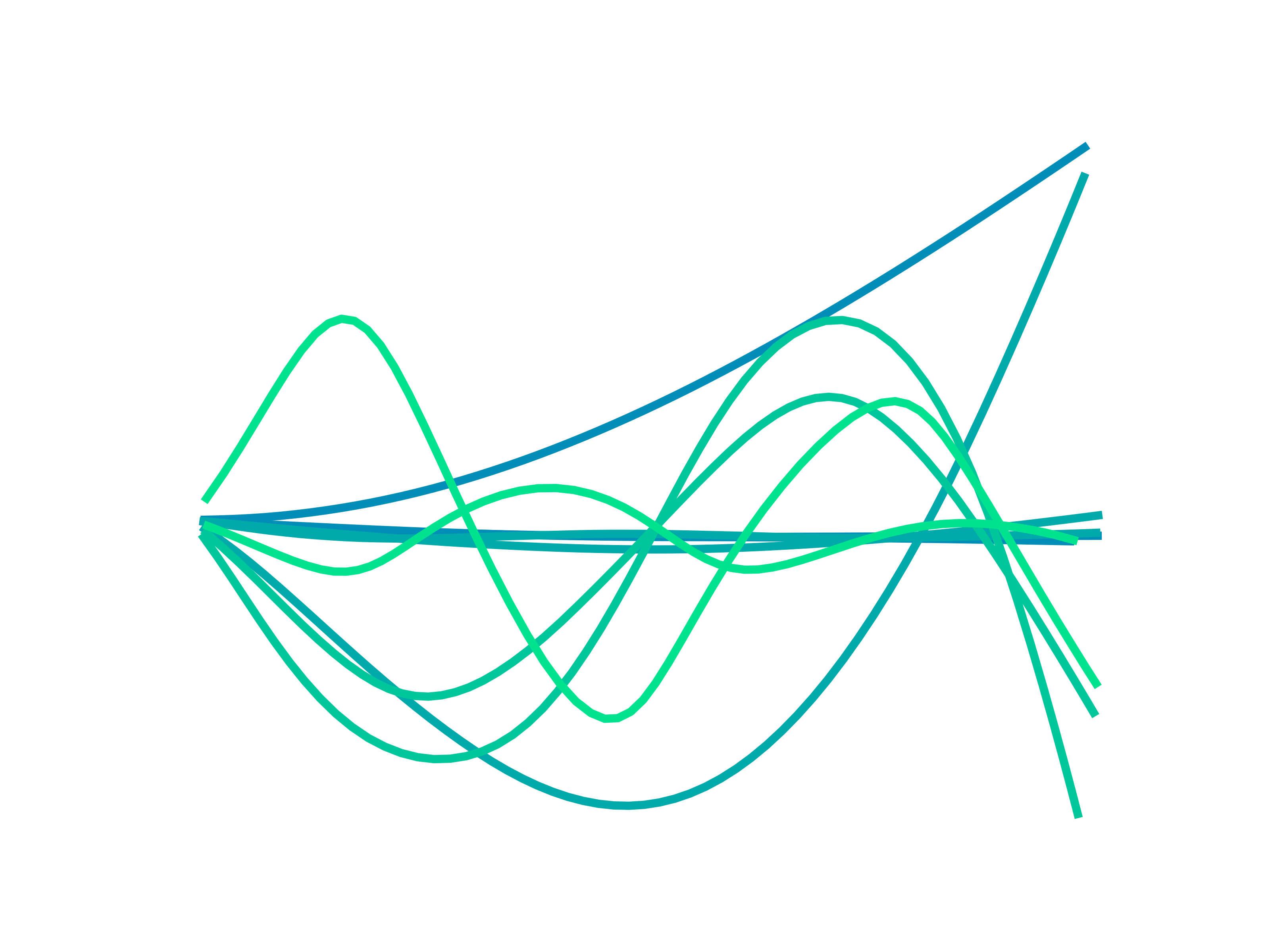}
            \caption{Modes from PCA.}
            \label{fig:lm lm}
    \end{subfigure}%
    \hfill
    \begin{subfigure}{0.29\textwidth}
            \includegraphics[width=\linewidth]{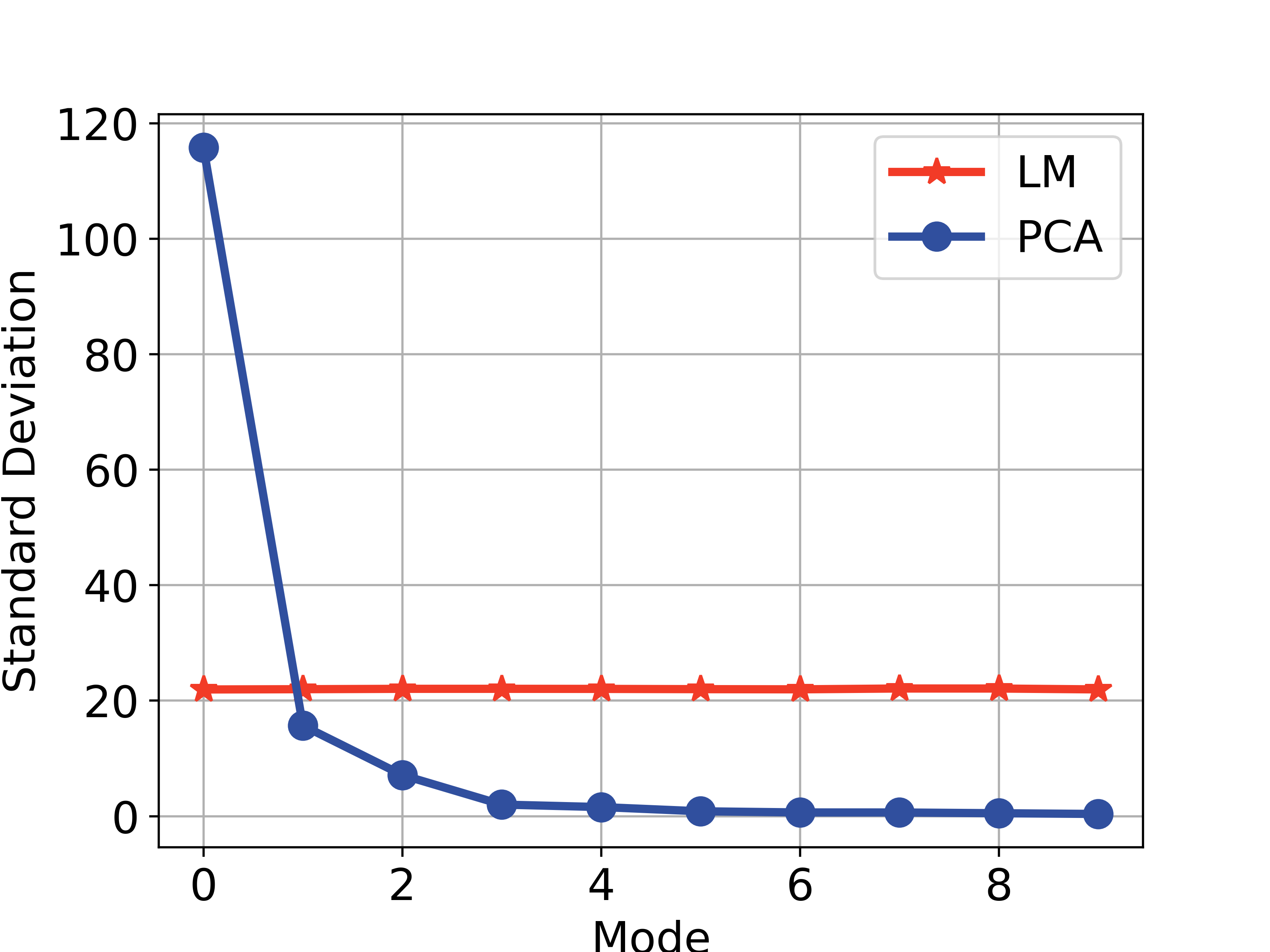}
            \caption{Variances of latents from PCA and LM.}
            \label{fig:std pca lm}
    \end{subfigure}%
    \caption{Analysis on PCA and Linear Mapping (Ours).}
    \label{fig: comparison on pca and lm}
\end{figure*}

To further improve the efficiency during the inference, we need to find a mapping $\mathcal{G}(\cdot)$ as simple as possible. \cite{jiang2023motiondiffuser} shows that PCA can have relatively low reconstruction error, and it is only a linear mapping which is computational efficient. Inspired by this, we use two linear mapping $U \in \mathbb{R}^{X \times Z}$ and $V \in \mathbb{R}^{Z \times X}$ to formulate the mapping function, $\mathcal{G}(\mbx)=\mbx U$ and $\mathcal{F}(\mbz)=\mbz V$. Since it is linear mapping, we call this method as Linear Mapping (LM). We set $Z=10$ and it works well both in reconstruction and performance of diffusion model. To satisfy distance-preserving requirements, columns in $V$ should be orthogonal to each other so each dimension in $\mbz$ is independent with each other. Thus, if this mapping is well-trained, variance of $\mbz$ is diagonal matrix. In our implementation, we assume we can learn such a good linear mapping and marginal variances in \cref{Corollary: ogd marginal only} are all diagonal matrices. We visualize the Top-10 components of PCA \cite{jiang2023motiondiffuser} and 10 components in $V$ in \cref{fig: comparison on pca and lm}. The modes learned from the LM exhibit greater diversity and expressiveness. In addition, latent variances derived from the LM across different dimensions are consistent, whereas PCA-derived latent variances show extreme variability, with a large variance in the main component and significantly lower variance in the remaining components. We also compare the performance of latent diffusion model with PCA and LM in \cref{tb: pca vs lm}, using Optimal Gaussian Diffusion. Our linear mapping approach can improve the performance.

\begin{table*}[tb]
\caption{\footnotesize Quantitative results on the Argoverse 2 validation dataset.}
\label{tb: pca vs lm}
\begin{center} {

\begin{tabular}{ccccc}
\toprule
Latent &$\text{avgMinADE}_6^*$&$\text{avgMinFDE}_6^*$&$\text{avgMinADE}_{128}$&$\text{avgMinFDE}_{128}$\\
\midrule

PCA \cite{jiang2023motiondiffuser} & 0.60 & 1.33& 0.46  & 0.80 \\
LM (Ours) & 0.60 & 1.32 & 0.43 & 0.81 \\
\bottomrule
\end{tabular}}
\end{center}
\end{table*}

\subsection{Sample Clustering Algorithm} \label{appendix:cluster}
We design a sample clustering algorithm to to generate a representative set from a batch of joint trajectory samples. Specifically, we first designate $\mathcal{U}$ as the central point for the groups. Then for each vehicle $i$, we map the trajectory samples $\mbx_{0,i}$, $i=1,2,...,N$ into closest $\mbr_i^l$, $l=1,2,...,L$. Thus, we assign each joint trajectory into $\mathcal{U}$. Subsequently, we sort the $\mathcal{U}$ by the size of its group memberships. Third, we prune the group based on the center of the groups represented by the combination of reference trajectories $\mbr_i^l$. Specifically, if the centers of two groups are close to each other, the group with the larger number of members will absorb the other. We define 'close enough' as a scenario where the maximum endpoint deviation between these two joint trajectories is less than 2.5 meters, aligning with the threshold used in NMS~\cite{shi2023mtr++}. After pruning, we calculate the probability of each group by its size. Since our clustering algorithm does not need to calculate the exact log probability \cite{jiang2023motiondiffuser}, it is less computationally demanding compared to more intensive techniques such as Non-Maximum Suppression (NMS) \cite{shi2023mtr++} or Expectation-Maximization (EM) \cite{varadarajan2022multipath++}. However, since we do not explicitly to calculate the likelihood of a sample, the likelihood estimation is not very accurate. This has resulted in our outperforming the rank-3rd benchmark on the Argoverse 2 Multi-world Forecasting leaderboard across all metrics, with the exception of $\text{avgBrierMinFDE}_K$, which is calculated based on the likelihood of the samples. Since we are focusing on optimizing diffusion model, we leave fast and accurate likelihood estimation for the future research. \Cref{tb: mtp leaderboard tune number of samples} demonstrates the influence of the number of samples. With more samples, all metrics improve. The performance saturates when the number of samples exceeds 1024.

\begin{table*}[tb]
\caption{\footnotesize Evaluation on different number of samples on the Argoverse 2 Multi-world Forecasting leaderboard. }
\label{tb: mtp leaderboard tune number of samples}
\begin{center} {
\resizebox{1.\textwidth}{!}{
\begin{tabular}{cccccccc}
\toprule
\# of Samples&$\text{avgMinFDE}_6^*$&$\text{avgMinFDE}_1^*$ &$\text{actorMR}_6^*$&$\text{avgMinADE}_6^*$&$\text{avgMinADE}_1^*$&$\text{avgBrierMinFDE}_6^*$&$\text{actorCR}_6^*$\\
\midrule
32 &1.34 & 2.78& 0.18 & 0.61 & 1.11 & 2.01 & 0.01 \\
128 &1.32 & 2.74& 0.17 & 0.60 & 1.09 & 1.97 & 0.01 \\
512 &1.31 & 2.72& 0.17 & 0.60 & 1.08 & 1.96 & 0.01 \\
1024 &1.31 & 2.71& 0.17 & 0.60 & 1.08 & 1.95 & 0.01 \\
2048 & 1.31 & 2.71& 0.17 & 0.60 & 1.08 & 1.95 & 0.01 \\
\bottomrule
\end{tabular}} }
\end{center}
\end{table*}

\subsection{Ablation Study on Marginal Predictors}
We froze the trained Optimal Gaussian Diffusion (OGD) model used in \cref{sec: exp jtp} and replaced QCNet\cite{zhou2023query} with Forecast-MAE \cite{cheng2023forecast} as the marginal predictor. We downloaded their pre-trained weights from their official website \footnote{https://github.com/jchengai/forecast-mae} and fine-tuned them for multiple epochs. Fine-tuning is necessary because the pre-trained weights were trained on a subset of target agents for the joint prediction task \cite{wilson2023argoverse}. We compare the marginal and joint metrics in \cref{tab: Comparison on different marginal predictors.}. The Forecast-MAE version of our model also yielded good performance, showing the flexibility of our approach. Note that the joint metrics of OGD are correlated with the marginal metrics. It indicates that we could improve OGD's performance without re-training if a better marginal predictor is available.

\begin{table*}
    \caption{Comparison on different marginal predictors.}
    \label{tab: Comparison on different marginal predictors.}
    \centering
  \begin{tabular}{cccccc}
\toprule
\multirow{2}*{Predictor}& \multirow{2}*{Epoch}&\multicolumn{2}{c}{Marginal Metrics}&\multicolumn{2}{c}{Joint Metrics}\\
 \multicolumn{1}{c}{}&\multicolumn{1}{c}{}& \multicolumn{1}{c}{$\text{MinADE}_{6}$} & \multicolumn{1}{c}{$\text{MinFDE}_{6}$} & \multicolumn{1}{c}{$\text{avgMinADE}_{128}$} & \multicolumn{1}{c}{$\text{avgMinFDE}_{128}$}\\

\midrule
QCNet&-&\textbf{0.37}&\textbf{0.60}&\textbf{0.43}&\textbf{0.81}\\
Forecast-mae&7&0.37&0.70&0.47&0.85\\
Forecast-mae&1&0.39&0.74&0.49&0.91\\
\bottomrule
\end{tabular} 
\end{table*}

\section{Guided Sampling Baseline and Step Size Tuning}
\label{app: guided sampling baseline and step size}
In this section, we first derive two guided sampling baselines: Next Noisy Mean (NNM) Guidance and Score Function (SF) Guidance. We also introduce the comparison settings for controllable generation tasks. Secondly, we present the step size tuning experiments.

\subsection{Guided Sampling Baseline}
Previous guided sampling approaches for controllable generation in autonomous driving are mainly focusing on guide the sample generation with an analytical guidance cost function $\cJ(\cdot)$ defined on clean data $\mbx_0$, such as goal point guidance and target speed guidance \cite{zhong2023guided}. In this paper, we discuss guided sampling approach under the human-defined guidance cost function $\cJ(\cdot)$ with no need to train additional guidance function defined on noisy data $\mbx_t$, $t>0$ \cite{janner2022planning}. 

The first baseline is to directly calculate $\nabla_{\mbx_t} \cJ(\mbx_t)$. Specifically, it first calculate the mean of $\mbx_{t-1}$ conditioned on $\mbx_t$, 
\begin{equation}
    \mbm_{t-1}=\frac{1}{\sqrt{\alpha_t}}(\mbx_t-\frac{\beta_t}{\sqrt{1-\Bar{\alpha}_t}}\mbepsilon_\theta(\mbx_t,t))
\end{equation}
Then, add the guidance into $\mbm_{t-1}$
\begin{equation}
    \label{eq: ecm nnm}
    \Tilde{\mbm}_{t-1}=\mbm_{t-1}-\text{clip}(\zeta \nabla_{\mbm_{t-1}}\cJ(\mbm_{t-1}),\pm \beta_t \mathbf{\sigma}_p)
\end{equation}
where $\zeta$ is the step size, $\text{clip}$ is elementwise clipping function, $\mathbf{\sigma}_p$ is the positive squared root of $\text{diag}[\mbSigma_p]$. Note that $\mbm_{t-1}$ is also noisy so $\cJ(\mbm_{t-1})$ suffers from numerical instability. Since this method directly inject the guidance into the mean of next noisy data distribution $\mbx_{t-1}$, we call it as Next Noisy Mean (NNM) Guidance.

The second baseline is to first project $\mbx_t$ into $\hat{\mbx}_0$ on the clean manifold and then calculate the guidance $\nabla_{\mbx_t}\cJ(\hat{\mbx}_0)$ \cite{rempe2023trace, jiang2023motiondiffuser}. Note that \cite{rempe2023trace, jiang2023motiondiffuser} directly train a denoiser to predict $\hat{\mbx}_0$. In our DDPM~\cite{ho2020denoising} formulation, $\hat{\mbx}_0$ is estimated through Tweedie’s formula,
\begin{equation}
    \hat{\mbx}_0 = \frac{1}{\sqrt{\Bar{\alpha}_{t}}}(\mbx_{t}-\sqrt{1-\Bar{\alpha}_{t}} \mbepsilon_\theta(\mbx_{t},{t}))
\end{equation}
Then, add the guidance $\nabla_{\mbx_t}\cJ(\hat{\mbx}_0)$ to the score function,
\begin{equation}
    \Tilde{\mbs}_\theta(\mbx_t,t) = \mbs_\theta(\mbx_t,t) + \text{clip}(\zeta \sqrt{1-\Bar{\alpha}_t}\nabla_{\mbx_t}\cJ(\hat{\mbx}_0),\pm  \mathbf{\sigma}_p)/\sqrt{1-\Bar{\alpha}_t}
\end{equation}
where $\zeta$ is the step size. Since $\mbepsilon_\theta(\mbx_t,t) =-\sqrt{1-\Bar{\alpha}_t} \mbs_\theta(\mbx_t,t)$, we have
\begin{equation}
    \Tilde{\mbepsilon}_\theta(\mbx_t,t) = \mbepsilon_\theta(\mbx_t,t) - \text{clip}(\zeta \sqrt{1-\Bar{\alpha}_t}\nabla_{\mbx_t}\cJ(\hat{\mbx}_0),\pm  \mathbf{\sigma}_p)
\end{equation}

We use DDIM \cite{song2020denoising} to accelerate the inference for controllable generation. Note that we replace $\beta_t$ in \cref{eq: ecm nnm} with $\sqrt{1-\Bar{\alpha}_t/\Bar{\alpha}_{t'}}$ for time index $t'$, $t' < t$.

To fairly compare the performance and efficiency of different guided sampling approaches, we set one network inference step followed by one gradient guidance step. For our ECM and ECMR, we set $t_k$ to be the same as the DDIM time step $t$. Specifically, we set DDIM step stride to 10 and $T$ to 100. Then $t_k=10(k+1)$, where $k={0,1,...,9}$.

\begin{figure*}[tb]
    \centering
    \begin{subfigure}[b]{0.33\textwidth}
            \includegraphics[width=\linewidth]{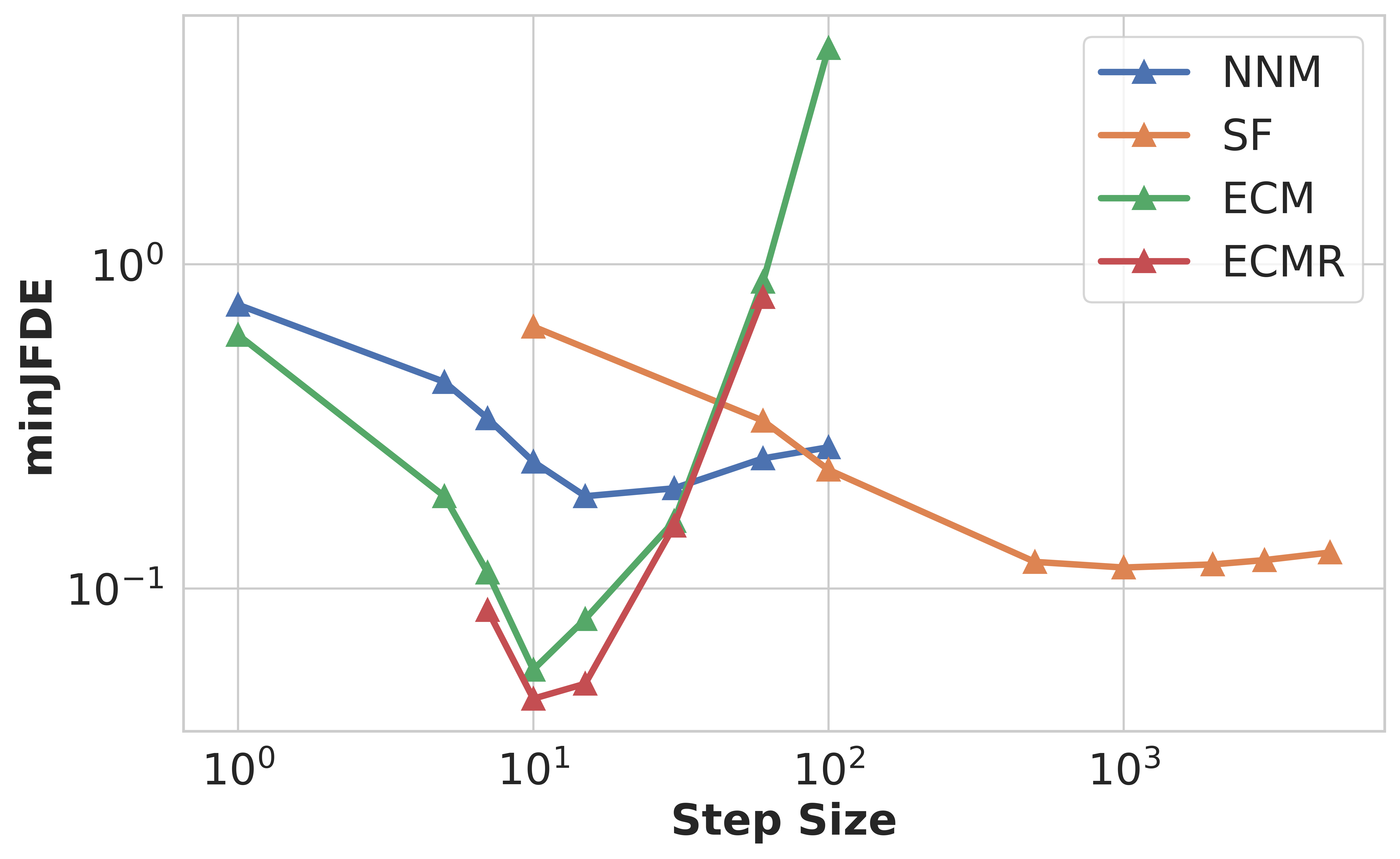}
            \caption{\textbf{GT}+\textbf{Normal Speed}}
            \label{fig:step size gt+n}
    \end{subfigure}%
    \hfill
    \begin{subfigure}[b]{0.33\textwidth}
            \includegraphics[width=\linewidth]{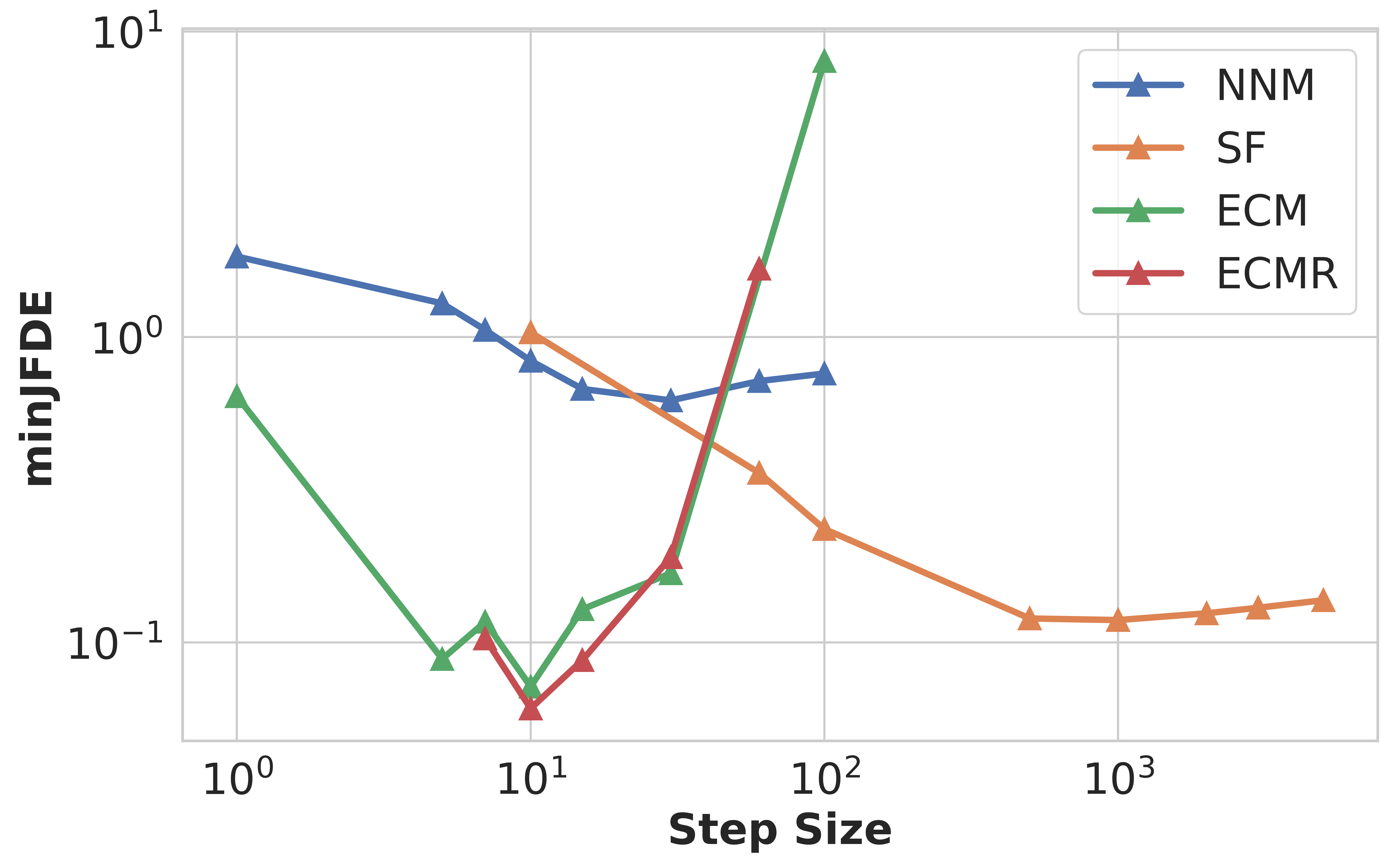}
            \caption{\textbf{GT}+\textbf{Deceleration}}
            \label{fig:step size gt+d}
    \end{subfigure}%
    \hfill
    \begin{subfigure}[b]{0.33\textwidth}
            \includegraphics[width=\linewidth]{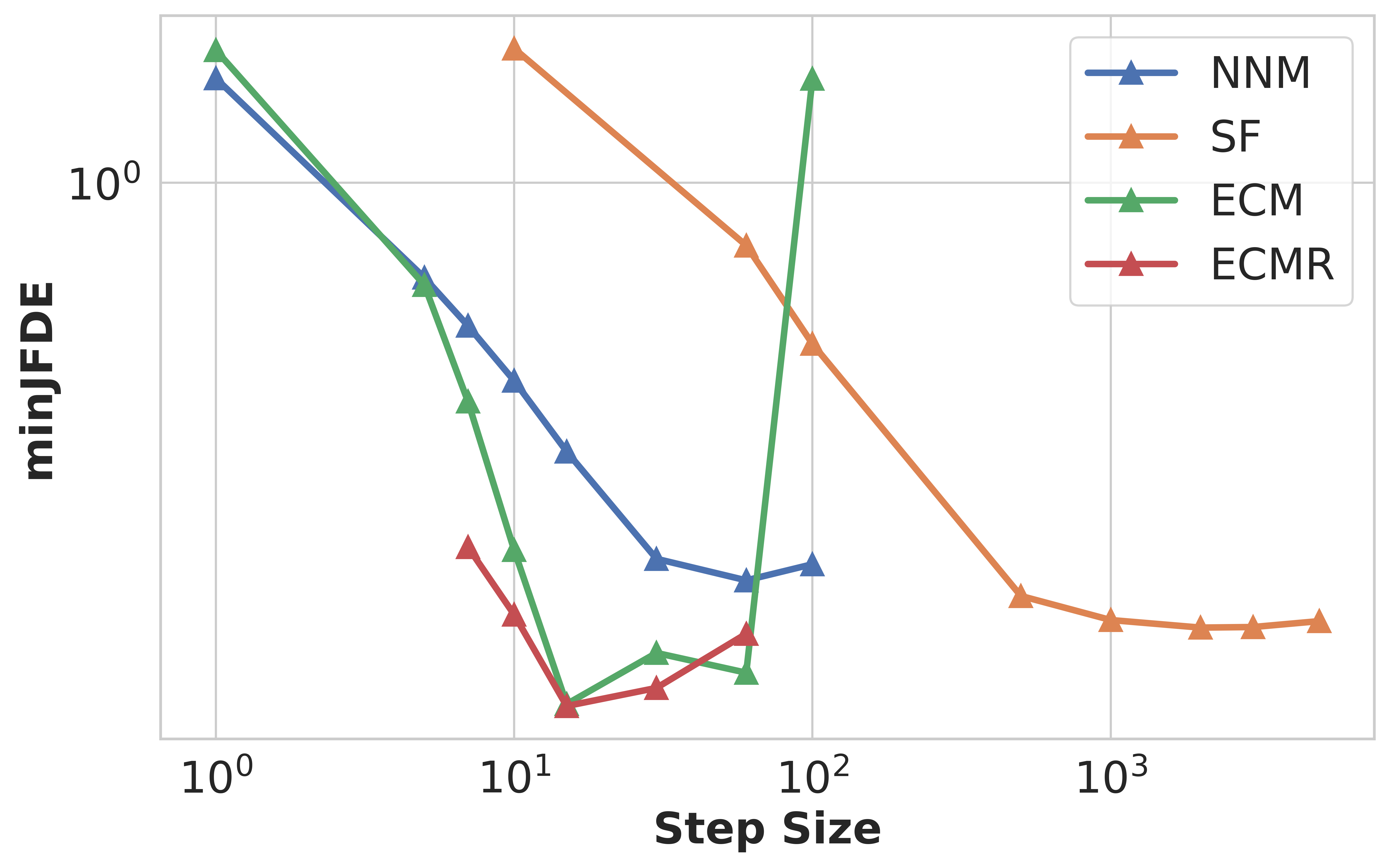}
            \caption{\textbf{GT}+\textbf{Acceleration}}
            \label{fig:step size gt+a}
    \end{subfigure}%
    \hfill
    \begin{subfigure}[b]{0.33\textwidth}
            \includegraphics[width=\linewidth]{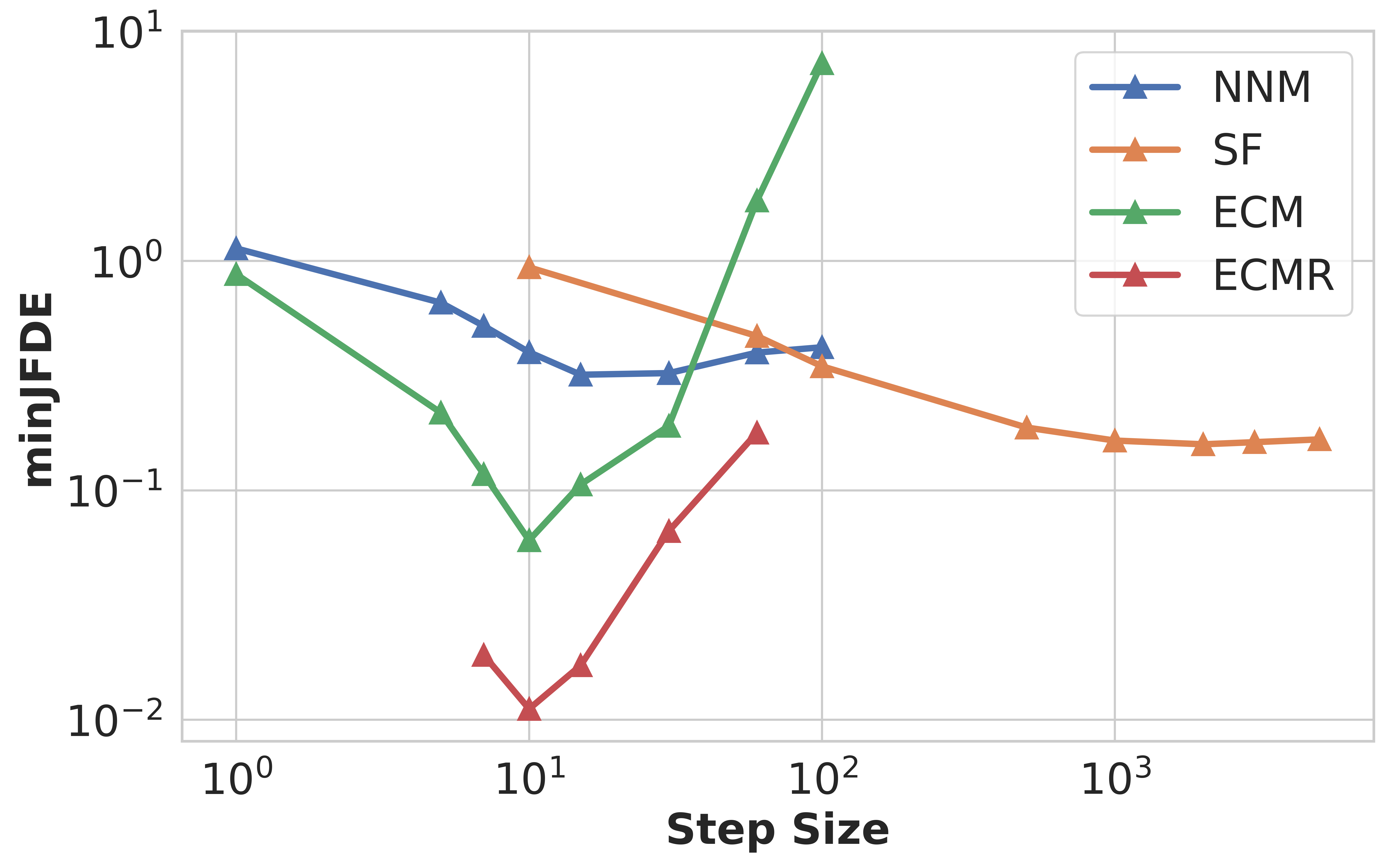}
            \caption{\textbf{U}+\textbf{Normal Speed}}
            \label{fig:step size u+n}
    \end{subfigure}%
    \hfill
    \begin{subfigure}[b]{0.33\textwidth}
            \includegraphics[width=\linewidth]{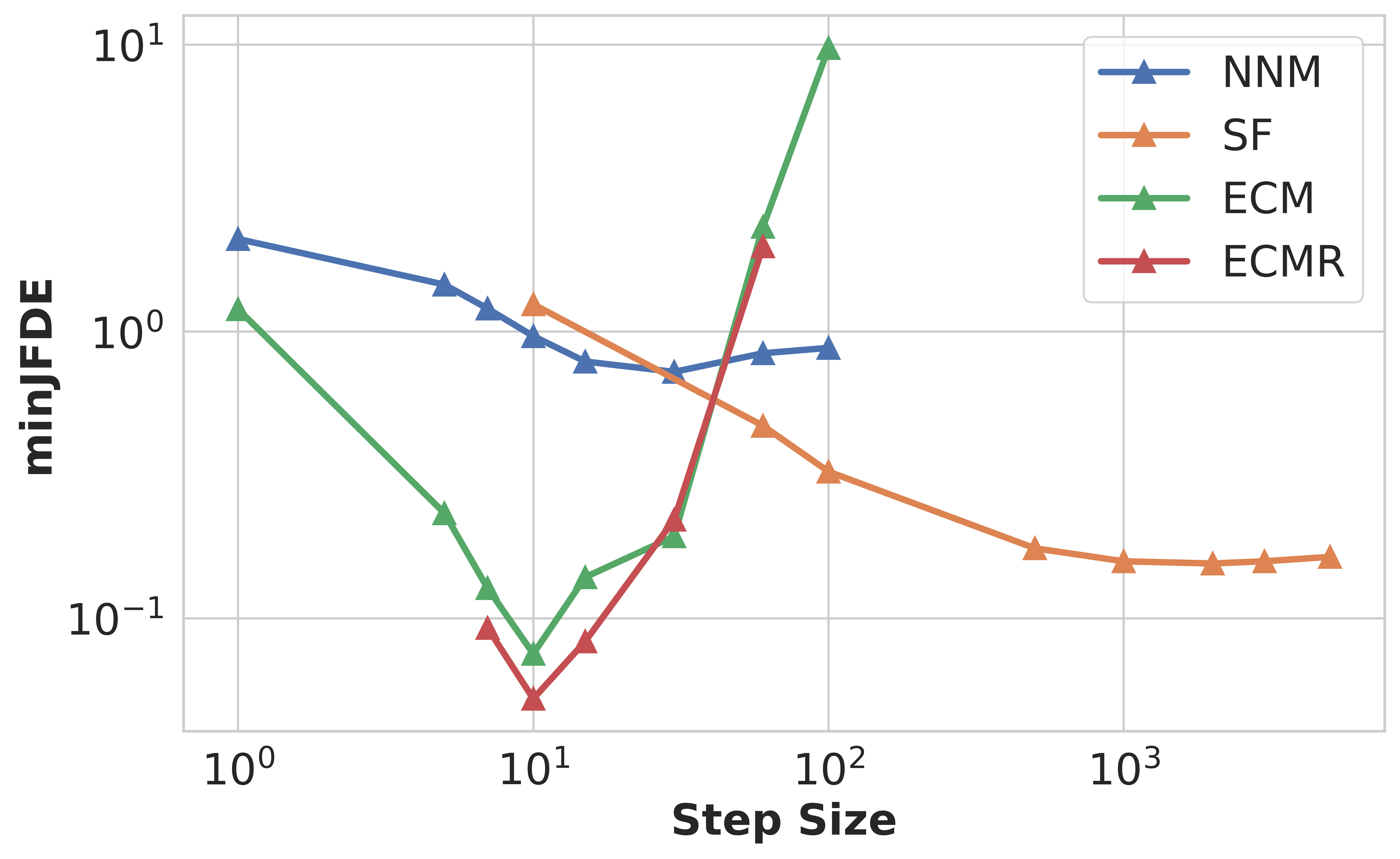}
            \caption{\textbf{U}+\textbf{Deceleration}}
            \label{fig:step size u+d}
    \end{subfigure}%
    \hfill
    \begin{subfigure}[b]{0.33\textwidth}
            \includegraphics[width=\linewidth]{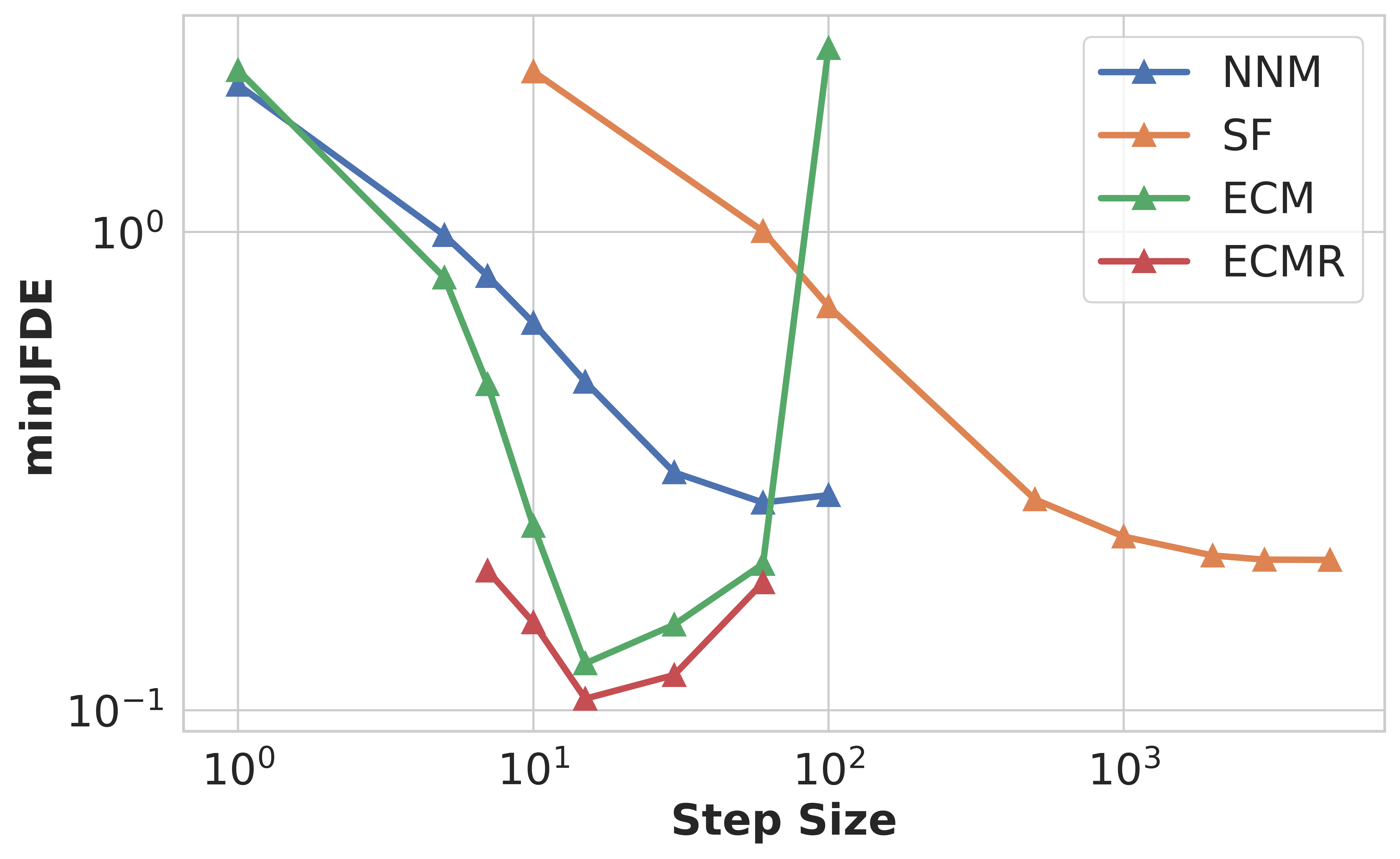}
            \caption{\textbf{U}+\textbf{Acceleration}}
            \label{fig:step size u+a}
    \end{subfigure}%
    
    \caption{Evaluation on the impact of step size. In general, small step size causes low guidance effectiveness. Big step size will cause the optimization process unstable, leading to low guidance effectiveness.}
    \label{fig: step size tuning}
\end{figure*}

\subsection{Step Size Tuning}
\label{app: step size tuning}

We conduct a grid search to identify the optimal gradient step size for NNM, SF, ECM, and ECMR. For NNM, ECM and ECMR, we use the settings $\zeta \in \{1,5,7,10,15,30,60,100\}$. For SF, $\zeta \in \{10,500,1000,2000,3000,5000\}$. We find that both excessively small and large gradient step sizes result in low guidance effectiveness, similar to the inefficiency or instability seen in gradient-based optimization algorithms with overly small or large step sizes. The optimal step size lies in between. Based on \cref{fig: step size tuning}, we select the optimal step size that yields the lowest minJFDE. We apply these optimal step sizes across all guided sampling methods and controllable generation tasks, and report the quantitative results using these sizes throughout the remainder of the paper.

\begin{figure*}[tb]
    \centering
    \begin{subfigure}{0.44\textwidth}
            \includegraphics[width=\linewidth]{images/minJFDE_all_tasks.png}
            \caption{minJFDE}
            \label{fig:minJFDE_all_tasks}
    \end{subfigure}%
    \hfill
    \begin{subfigure}{0.44\textwidth}
            \includegraphics[width=\linewidth]{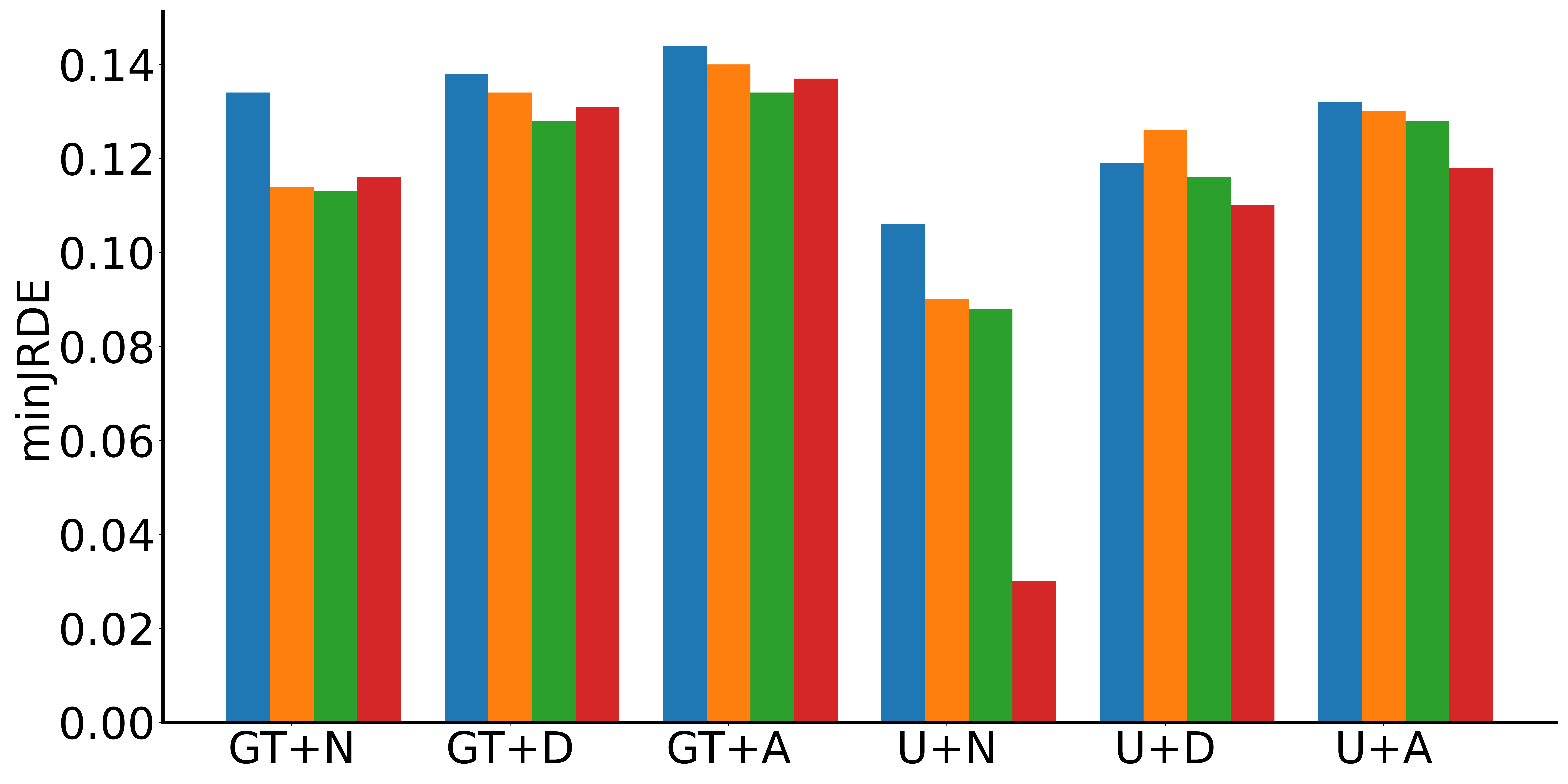}
            \caption{minJRDE}
            \label{fig:minJRDE_all_tasks}
    \end{subfigure}%
    \hfill
    \begin{subfigure}{0.44\textwidth}
            \includegraphics[width=\linewidth]{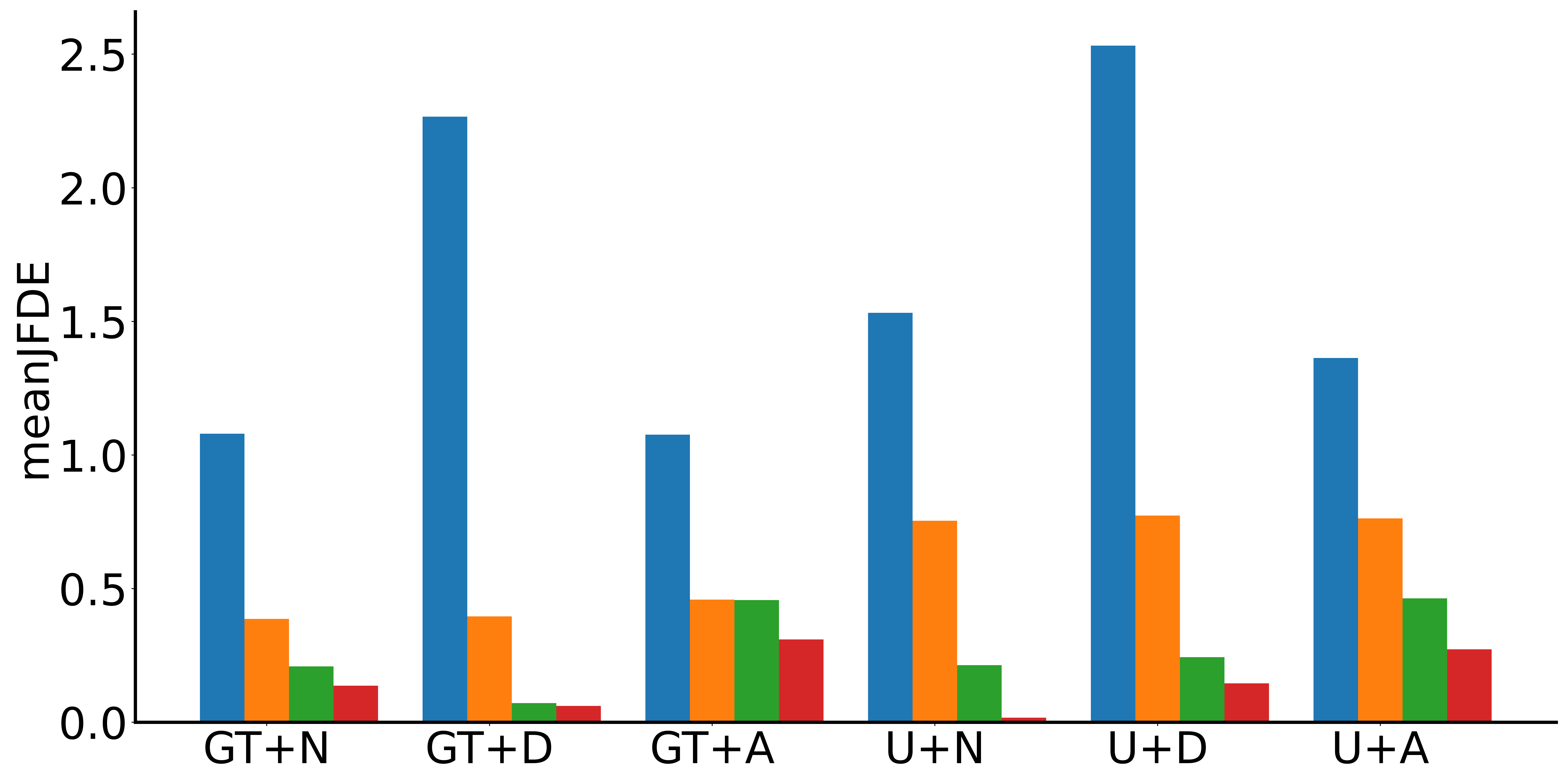}
            \caption{meanJFDE}
            \label{fig:meanJFDE_all_tasks}
    \end{subfigure}%
    \hfill
    \begin{subfigure}{0.44\textwidth}
            \includegraphics[width=\linewidth]{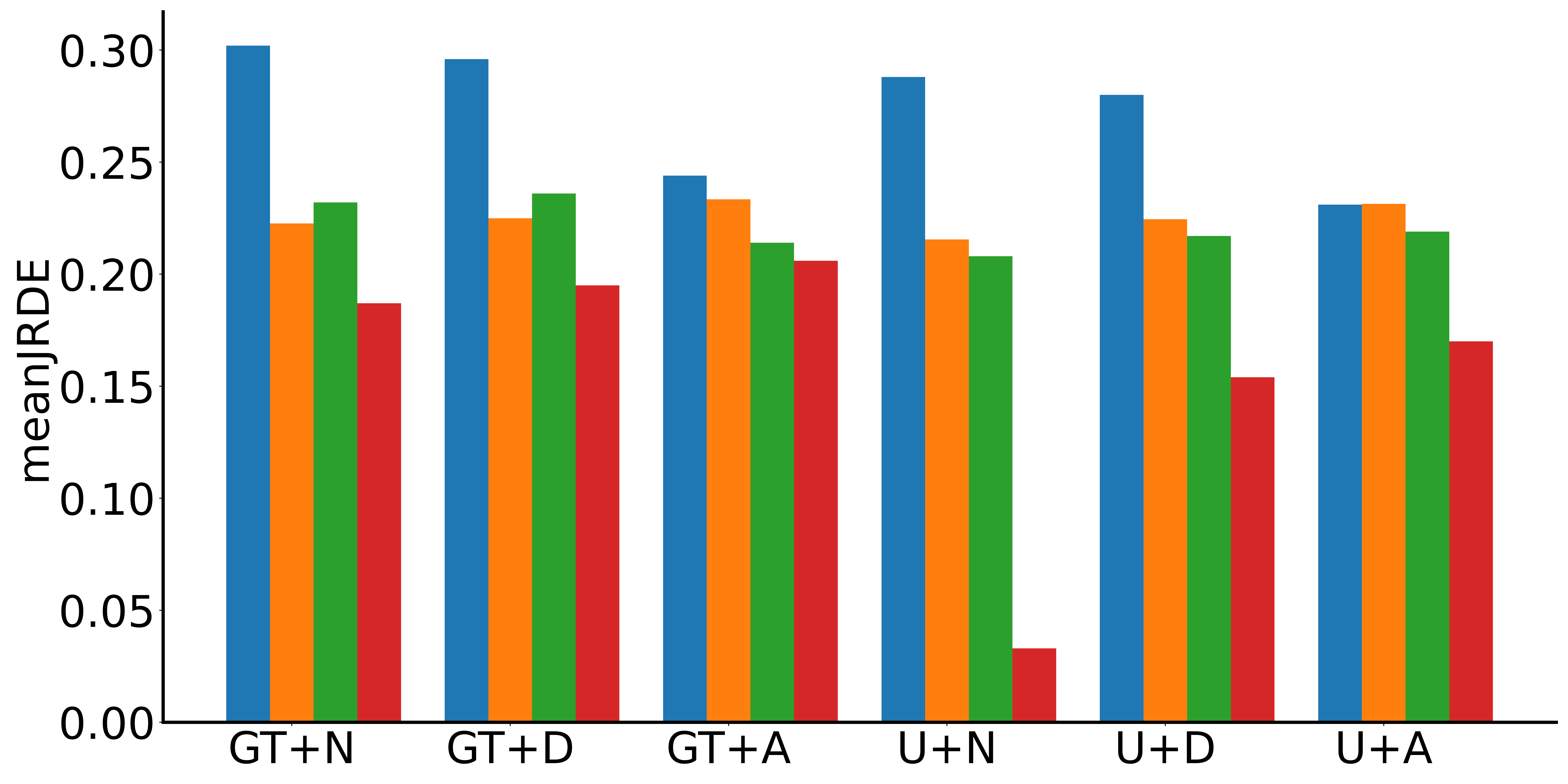}
            \caption{meanJRDE}
            \label{fig:meanJRDE_all_tasks}
    \end{subfigure}%
    \caption{Evaluation on different guided sampling methods.}
    \label{fig: controllable generation eval all}
\end{figure*}

\section{Evaluation on Controllable Generation Tasks}
\label{app: eval on controllable generation task }
\Cref{fig: controllable generation eval all} showcases the performance of various guided sampling methods across different tasks. ECM and ECMR consistently outperform others in minJFDE across all tasks and in minJRDE in nearly all tasks, indicating their superior guidance effectiveness and realism. Notably, ECMR, in particular, excels in meanJFDE/meanJRDE, suggesting the generated joint trajectories are of higher average quality. This implies our methods' ability to produce valid and high-quality joint trajectories with fewer samples. The experiment results are shown in \cref{tb: guided sampling number of samples}. Our ECM and ECMR approaches can generate a better joint trajectory with a small number of samples.

\begin{figure*}[t]
    \centering
    \includegraphics[width=1.0\linewidth]{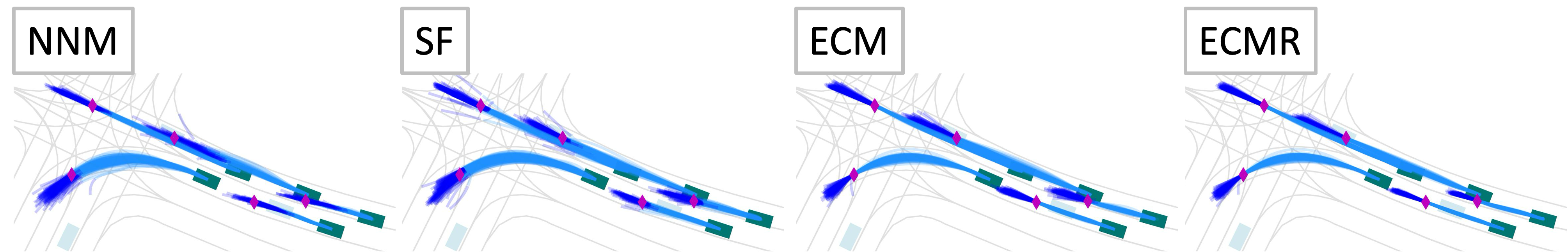}
    \caption{Visualizations on controllable generation: route set \textbf{GT} and \textbf{Acceleration}. Magenta diamonds represent goal points. Dodgerblue curves represent the predicted joint trajectory from 0s to 5s. Blue curves represent the predicted joint trajectory from 5s to 6s.}
    \label{fig:goal_at5s_visual}
\end{figure*}

\begin{figure*}[t]
    \centering
    \includegraphics[width=1.0\linewidth]{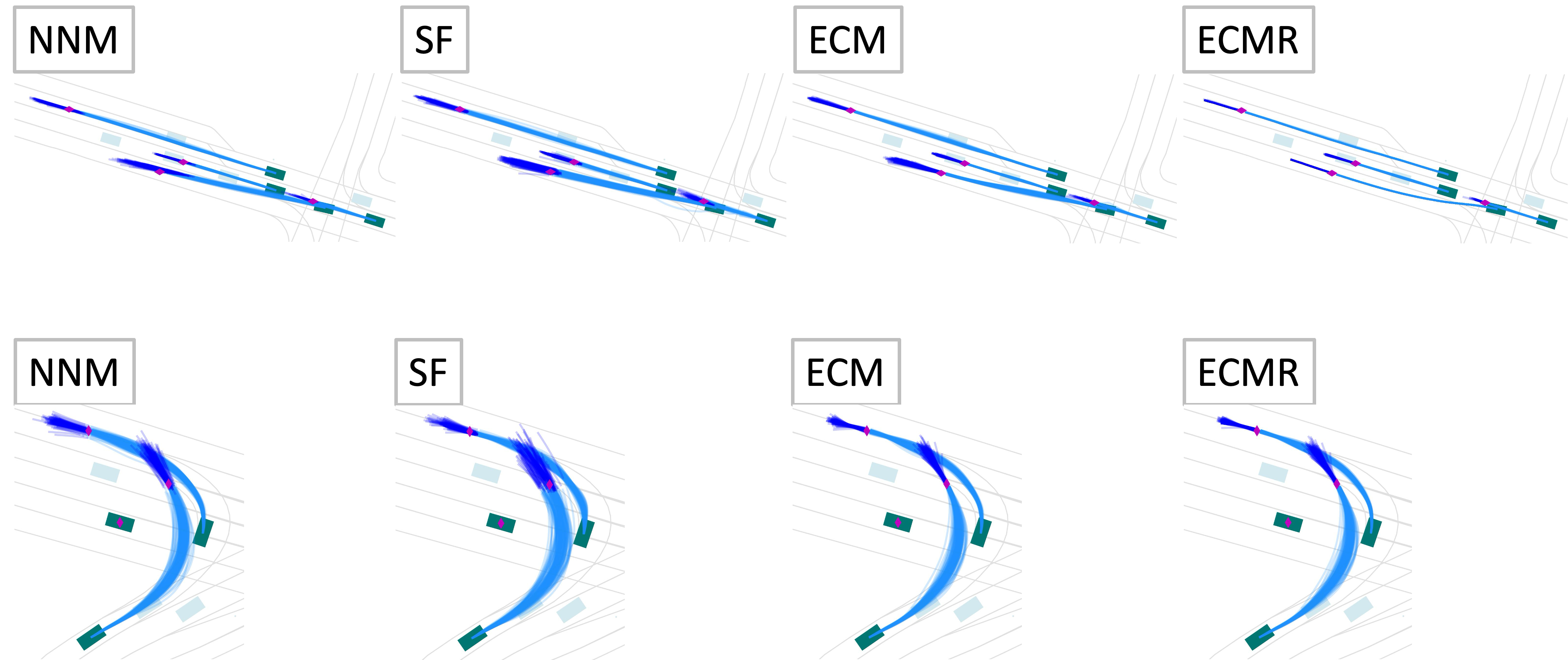}
    \caption{Visualizations on controllable generation: route set \textbf{U} and \textbf{Acceleration}. Magenta diamonds represent goal points. Dodgerblue curves represent the predicted joint trajectory from 0s to 5s. Blue curves represent the predicted joint trajectory from 5s to 6s.}
    \label{fig:rand_goal_at5s_visual}
\end{figure*}

\Cref{fig:goal_at5s_visual} and \cref{fig:rand_goal_at5s_visual} demonstrate examples in different controllable generation tasks. Our guided sampling approach, ECM, can better satisfy the guidance while simultaneously maintaining realism. With reference joint trajectory, ECM can further enhance its performance.

\begin{table*}[tb]
\caption{\footnotesize Evaluation on different number of samples in the controllable generation task: route set \textbf{GT} and \textbf{Normal Speed}.}
\label{tb: guided sampling number of samples}
\begin{center} {

\begin{tabular}{cccc}
\toprule
Sampling  &\# of samples &minJFDE&minJRDE\\
\midrule

NNM \cite{zhong2023guided} & 128 & 0.193& 0.134 \\
SF \cite{rempe2023trace, jiang2023motiondiffuser} & 128 & 0.116 & 0.114 \\
\midrule
ECM & 128 & 0.056& 0.113 \\
ECM & 96 & 0.068& 0.113 \\
ECM & 64 & 0.135& 0.118 \\
ECM & 32 & 0.351& 0.143 \\
\midrule
ECMR & 128 & 0.046 & 0.115 \\
ECMR & 96 & 0.046 & 0.116 \\
ECMR & 64 & 0.088 & 0.119 \\
ECMR & 32 & 0.368 & 0.993 \\

\bottomrule
\end{tabular}}
\end{center}
\end{table*}

\end{document}